\pdfoutput=1
\documentclass[11pt]{article}
\usepackage[dvipsnames]{xcolor}

\usepackage[final]{acl}

\usepackage{times}
\usepackage{latexsym}

\usepackage[T1]{fontenc}
\usepackage[utf8]{inputenc}
\usepackage{microtype}

\usepackage{inconsolata}

\usepackage{graphicx}
\usepackage{amsmath}
\usepackage{enumitem}
\usepackage{booktabs}
\usepackage{enumerate}
\usepackage{amssymb}
\usepackage{amsthm}
\usepackage{pifont}
\usepackage{multirow}
\usepackage{multicol}
\usepackage{cleveref}
\usepackage{tabularx}
\usepackage{CJKutf8}
\usepackage{xcolor,colortbl}

\definecolor{Gray}{gray}{0.90}
\definecolor{LightCyan}{rgb}{0.88,1,1}

\definecolor{purp}{HTML}{791f87}

\newcolumntype{a}{>{\columncolor{Gray}}c}
\mathchardef\mhyphen="2D % Define a "math hyphen"
\newcommand{\Sref}[1]{\S\ref{#1}}

\title{Teaching LLMs to Abstain across Languages via Multilingual Feedback}

\author{Shangbin Feng\textsuperscript{1} \ \ Weijia Shi\textsuperscript{1} \ \ Yike Wang\textsuperscript{1} \ \ Wenxuan Ding\textsuperscript{2} \ \ Orevaoghene Ahia\textsuperscript{1} \\
\textbf{Shuyue Stella Li\textsuperscript{1}} \ \ \ \textbf{Vidhisha Balachandran\textsuperscript{3}} \ \ \ \textbf{Sunayana Sitaram\textsuperscript{3}} \ \ \ \textbf{Yulia Tsvetkov\textsuperscript{1}} \\
\textsuperscript{1}University of Washington \ \ \textsuperscript{2}The University of Texas at Austin \ \ \textsuperscript{3}Microsoft Research \\
\href{mailto:shangbin@cs.washington.edu}{\texttt{shangbin@cs.washington.edu}}
}

\begin{document}
\maketitle

\begin{abstract}
%Teaching large language models (LLMs) to abstain in low-confidence scenarios has become an important strategy to mitigate hallucinations and enhance reliability. While existing studies on LLM abstention are mostly English-only, abstaining beyond English is of critical importance given the diminishing factuality of LLMs on the long tail of languages.
Multilingual large language models (LLMs) often have knowledge disparities across languages, with larger gaps in under-resourced languages. % due to factors such as language resourceness.
Teaching LLMs to abstain in the face of knowledge gaps is thus a promising strategy to mitigate hallucinations in multilingual settings. However, previous studies on LLM abstention primarily focus on English; we find that directly applying these solutions beyond English results in up to 20.5\% performance gaps between high and low-resource languages, potentially due to LLMs' drop in calibration and reasoning beyond a few resource-rich languages. To this end, we propose strategies to enhance LLM abstention by \emph{learning from multilingual feedback}, where LLMs self-reflect on proposed answers in one language by generating multiple feedback items in related languages: we show that this helps identify the knowledge gaps across diverse languages, cultures, and communities. Extensive experiments demonstrate that our multilingual feedback approach outperforms various strong baselines, achieving up to 9.2\% improvement for low-resource languages across three black-box and open models on three datasets, featuring open-book, closed-book, and commonsense QA. Further analysis reveals that multilingual feedback is both an effective and a more equitable abstain strategy to serve diverse language speakers, and cultural factors have great impact on language selection and LLM abstention behavior, highlighting future directions for multilingual and multi-cultural reliable language modeling.\footnote{Code and data are publicly available at \href{https://github.com/BunsenFeng/M-AbstainQA}{https://github.com/BunsenFeng/M-AbstainQA}.}
\end{abstract}

\section{Introduction}
\label{sec:introduction}
Large language models (LLMs) encode extensive information and aid knowledge-intensive tasks \citep{petroni2019language, brown2020language, yu2023kola}. However, knowledge gaps and subsequent model hallucinations pose an everlasting challenge that compromises LLM reliability  \citep{lazaridou2021mind, ji2023survey, kumar2023language, mishra2024fine}. A growing body of work seeks to enhance LLM reliability by teaching them to \emph{abstain}, i.e., avoiding wrong answers in low-confidence scenarios to mitigate hallucinations and factual inaccuracies. While these studies put forward viable solutions, they are evaluated on English only \citep{gu-hopkins-2023-evaluation, varshney-baral-2023-post, yang2023alignment, feng2024don}. However, the factuality of multilingual LLMs in low-resource languages is often worse \citep{zhang-etal-2023-dont, lai2023okapi, kang2024comparing}, underserving diverse language speakers and communities. As such, there is an urgent need for robust abstaining strategies that work with the long tail of languages.

\begin{figure}
    \centering
    \includegraphics[width=1\linewidth]{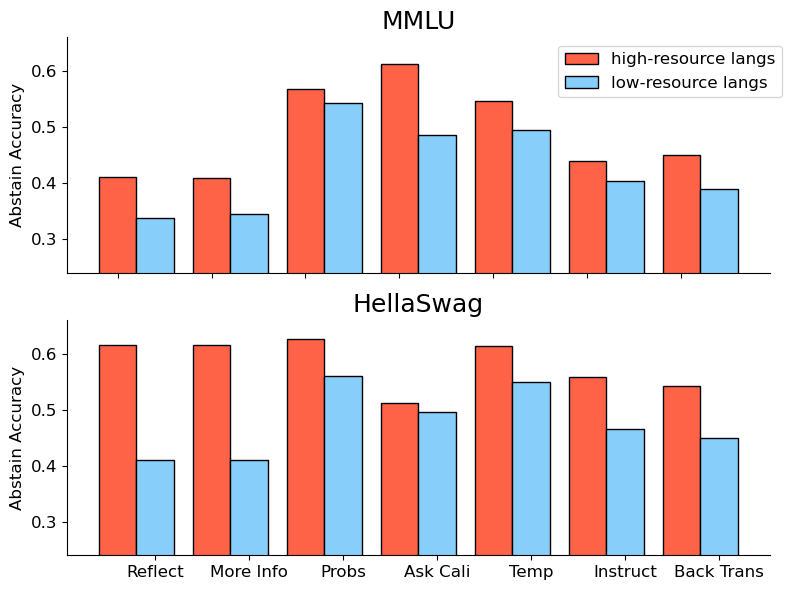}
    \caption{Average accuracy of abstention baselines in low- and high-resource languages with \textsc{Aya-13B}. \textbf{Existing abstain strategies drop by 8.4\% on average when applied to QA in low-resource languages.}}
    \vspace*{-15pt}
    \label{fig:teaser}
\end{figure}

Developed and evaluated in English, are existing abstain approaches viable for low-resource languages? Drawing from \citet{feng2024don}, we first evaluate seven existing abstain methods, spanning calibration \citep{jiang2021can, tian2023just}, prompting \citep{edunov2018understanding, kadavath2022language, Feng2023KnowledgeCF}, and training \citep{ouyang2022training}, on multilingual MMLU and Hellaswag datasets \citep{lai2023okapi} featuring 8 high-resource and 7 low-resource languages. Figure \ref{fig:teaser} demonstrates that performance degrades by up to 12.8\% and 20.5\% for both datasets: while existing approaches perform well in English, they are limited by LLMs' diminishing utility and calibration accuracy in low-resource languages, struggling to identify knowledge gaps and abstain accordingly. As a result, we ask: \emph{how to identify knowledge gaps in LLMs and reliably abstain beyond English?}

To this end, we present the first study on multilingual LLM abstention and propose to teach LLMs to abstain by \emph{generating and learning from multilingual feedback in related languages} (Figure \ref{fig:overview}). While the concept of generated feedback was previously demonstrated to improve reasoning and alignment in English-only scenarios \citep{du2023improving, madaan2024self}, sampling diverse and high-quality feedback in low-resource languages is challenging due to LLMs' diminishing utility in long-tail languages \citep{lai2023okapi}. Multilingual LLMs can leverage related languages to improve performance via transfer learning \cite{lin-etal-2019-choosing, pires-etal-2019-multilingual, asai2023buffet, tanwar-etal-2023-multilingual}, so we expect that generating feedback from \emph{related languages} would help identify knowledge gaps across diverse domains and cultures. Therefore, we probe multilingual LLMs to provide feedback, on its proposed answer, in several \emph{related} languages, where language relatedness is defined by linguistic typology, geography, or culture \citep{littell2017uriel,lin2019choosing,sun2021cross}. Together with the proposed answer and generated feedback from the most related languages, LLMs reason and self-reflect to make abstain decisions.

%it might be challenging to sample diverse and high-quality feedback in one low-resource language \citep{lai2023okapi}. As a result, after QA in one language, we probe LLMs to provide feedback on its proposed answer in several \emph{related} languages, where language relatedness is defined by linguistic theories such as typology and phonology \citep{littell2017uriel}. We expect feedback from related languages to identify the knowledge gaps across diverse domains and cultures to provide a well-rounded assessment of the proposed answer. Together with the proposed answer and generated feedback from the most related languages, LLMs reason and self-reflect to make abstain decisions.

We evaluate baselines and our \emph{multilingual feedback} approach using three open-source and proprietary LLMs (\emph{GPT-4}, \emph{Aya-13B}, and \emph{ChatGPT}) on three datasets in open-domain, closed-book, and commonsense QA. Extensive experiments demonstrate that \emph{multilingual feedback} consistently outperforms strong baselines across models and datasets, achieving up to 9.2\% improvements of abstain accuracy for low-resource languages. Further analysis reveals that \emph{multilingual feedback} presents a more equitable abstain strategy, highlighting culture as a driving factor in multilingual abstention. It impacts the optimal languages for feedback and LLMs' performance gaps across diverse information domains.

\section{Methodology}
\label{sec:methodology}

\paragraph{Background} We focus on teaching LLMs to \textbf{Abstain} in \textbf{Q}uestion \textbf{A}nswering (AbstainQA) \citep{feng2024don}: given a query $\boldsymbol{q}$ and an $\mathrm{LLM}$, we aim to develop robust abstention strategies $f(\boldsymbol{q}, \mathrm{LLM}) \rightarrow \{\textit{true}, \textit{false}\}$. Ideally, the LLM abstains ($f = \textit{true}$) when it would provide an incorrect answer and should not abstain ($f = \textit{false}$) when it is capable of generating a correct answer \citep{feng2024don}. $f$ should work for diverse languages of varying language families, resourceness levels, and speaker communities.

%While existing English-only studies employed calibration \citep{jiang2021can, tian2023just}, prompting \citep{edunov2018understanding, kadavath2022language, Feng2023KnowledgeCF}, training \citep{ouyang2022training, azaria2023internal, cobbe2021training}, and consistency-based methods \citep{wang2022self, feng2024don} for AbstainQA, they mostly relied on generating and reasoning in English: Figure \ref{fig:teaser} demonstrates that these approaches often have substantial performance gaps between high and low-resource languages, suggesting that monolingual reasoning in the language of the question might not be as reliable in low-resource languages, potentially due to LLMs' diminishing utility beyond English. To this end, we propose to \emph{teach LLMs to abstain via multilingual feedback}, hypothesizing that self-feedback items about its proposed answer from related languages could help identify the blind spots across cultures, perspectives, and contexts. We present an overview in Figure \ref{fig:overview}.

\begin{figure}
    \centering
    \includegraphics[width=1\linewidth]{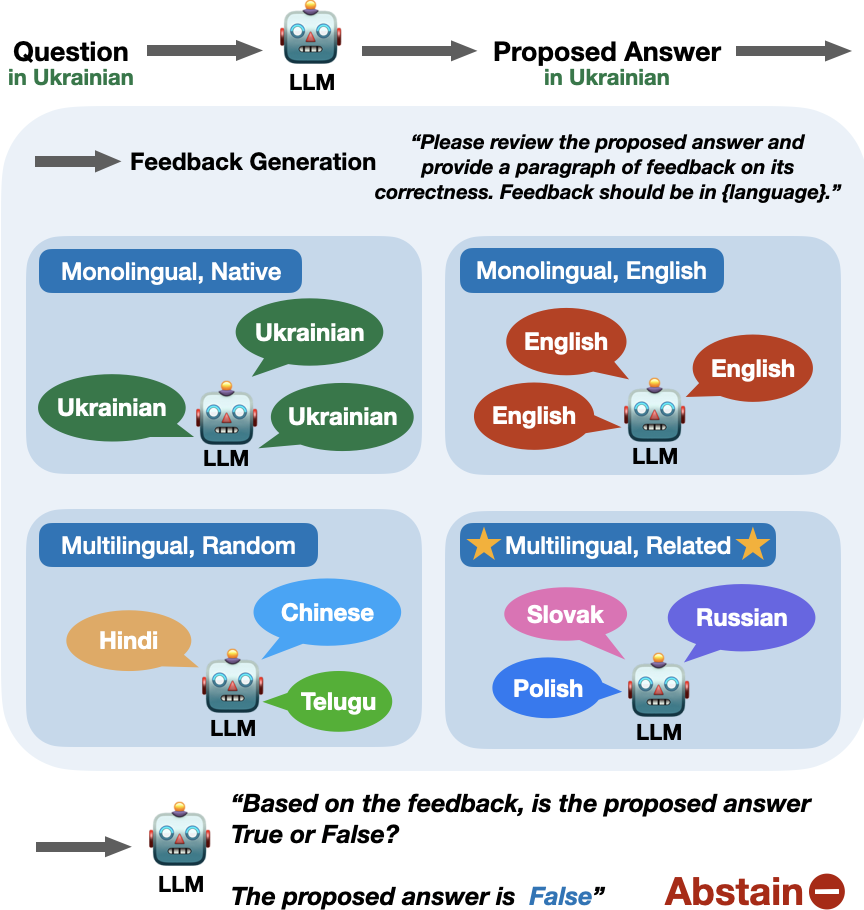}
    \caption{\textbf{Overview of \emph{abstaining via multilingual feedback}.} LLMs generate feedback on its proposed answer in four language settings to make abstain decisions.}
    %\vspace*{-15pt}
    \label{fig:overview}
\end{figure}

Since existing approaches to LLM abstention are limited by LLMs' diminishing utility and calibration beyond English (Figure \ref{fig:teaser}, \Sref{sec:introduction}), we propose to \emph{teach LLMs to abstain via multilingual feedback}, hypothesizing that self-feedback about its proposed answer from related languages could help identify the blind spots across cultures, perspectives, and contexts. We present an overview in Figure \ref{fig:overview}.

\paragraph{Abstain with Multilingual Feedback} LLMs take three steps to make an abstain decision:

\begin{itemize}[leftmargin=*]
    \item Given a question $\boldsymbol{q}$, the LLM first generates a proposed answer: $\boldsymbol{a} = \mathrm{LLM}(\boldsymbol{q})$.
    \item We probe the LLM itself for multilingual feedback on its proposed answer, concretely $\boldsymbol{f}_i = \mathrm{LLM}(\boldsymbol{q}, \boldsymbol{a} \mid \boldsymbol{\ell}_i)$ where feedback $\boldsymbol{f}_i$ is generated in language $\boldsymbol{\ell}_i$. We specifically use the prompt \textit{``Please review the proposed answer and provide a paragraph of feedback on its correctness. Feedback should be in $\boldsymbol{\ell}_i$.''} to elicit $\boldsymbol{f}_i$.
    \item Finally, the LLM employs the multilingual feedback to reason and make an abstain decision: $\mathrm{LLM(\boldsymbol{q}, \boldsymbol{a}, \{\boldsymbol{f}_1, \boldsymbol{f}_2, \cdots, \boldsymbol{f}_k\}) \rightarrow \{\textit{true}, \textit{false}\}}$. We specifically use the prompt \textit{``Based on the feedback, is the proposed answer True or False?''} and abstain if the answer $\boldsymbol{a}$ is deemed false.
\end{itemize}

\paragraph{Language Selection} Contrary to English-only scenarios, it is often challenging to sample diverse and high-quality feedback in one low-resource language. We hypothesize that by generating feedback in related languages to the language of the question $\boldsymbol{\ell}$, LLMs could better identify internal knowledge gaps and patch the blind spots with information across varying cultures, perspectives, and more. We experiment with four modes of selecting feedback languages $\{\boldsymbol{\ell}_1, \cdots, \boldsymbol{\ell}_k\}$.

\begin{itemize}[leftmargin=*]
    \item \underline{\emph{monolingual, native}} (\textsc{Mono-native}): all feedback are sampled in the native language of the question: $\boldsymbol{\ell}_1 = \cdots = \boldsymbol{\ell}_k = \boldsymbol{\ell}$. This resembles the previous English-only setting where questions and feedback are in the same language (English).
    \item \underline{\emph{monolingual, English}} (\textsc{Mono-English}): regardless of the language of the question, all feedback are sampled in English: $\boldsymbol{\ell}_1 = \cdots = \boldsymbol{\ell}_k =  \textit{English}$. This is because English is the highest-resource language and is often used as the source language in cross-lingual transfer  \citep{conneau2018xnli, conneau2019cross, hu2020xtreme,wang-etal-2020-negative}.
    \item \underline{\emph{multilingual, random}} (\textsc{Multi-random}): this is a control setting where we employ multiple languages for feedback generation, but the languages are randomly selected from a language pool $\mathcal{L}$: $\boldsymbol{\ell}_i = \mathrm{random\_choice}(\mathcal{L})$.
    \item \underline{\emph{multilingual, related}} (\textsc{Multi-related}): we propose to employ languages related to the language of the question $\boldsymbol{\ell}$ for feedback generation. Concretely, we employ Lang2vec \citep{littell2017uriel} to obtain the vector representation of a language $\boldsymbol{v}_{\boldsymbol{\ell}}^a$ in a linguistic attribute $a \in \mathcal{A}$.\footnote{Six attributes are considered in Lang2vec: syntactic, geographic, phonological, genetic, inventory, and featural.} We define the distance between a pair of languages as an average of distances across attributes:
\begin{align*}
    \mathrm{dist}(\boldsymbol{\ell}, \boldsymbol{\ell'}) = \frac{1}{|\mathcal{A}|} \sum_{a \in \mathcal{A}} \frac{{(\boldsymbol{v}_{\boldsymbol{\ell}}^{a})}^T \boldsymbol{v}_{\boldsymbol{\ell'}}^{a}}{\|\boldsymbol{v}_{\boldsymbol{\ell}}^{a}\| \| \boldsymbol{v}_{\boldsymbol{\ell'}}^{a}\|}
\end{align*}
The $k$ languages with the least distance to $\boldsymbol{\ell}$ are then selected for feedback generation: $\{\boldsymbol{\ell}_1, \cdots, \boldsymbol{\ell}_k\} = \mathrm{argmin \mhyphen k}_{\boldsymbol{\ell'}} \ \mathrm{dist}(\boldsymbol{\ell}, \boldsymbol{\ell'})$. We employ $k=3$ multilingual feedback by default.

\end{itemize}

\begin{table*}[t]
    \centering
    \setlength{\tabcolsep}{1pt}
    \renewcommand{\arraystretch}{0.8}
    \resizebox{1\textwidth}{!}{
    \begin{tabular}{laccccccca|accccccca}
    \toprule[1.5pt]
    \multirow{2}{*}{\textbf{Method}} & \multicolumn{9}{c|}{\textbf{M-MMLU}} & \multicolumn{9}{c}{\textbf{M-Hellaswag}} \\
     & Avg-H & bn & ta & ne & ml & mr & te & kn & Avg-L & Avg-H & bn & ta & ne & ml & mr & te & kn & Avg-L \\ \midrule[0.75pt]
      \multicolumn{19}{c}{\textbf{\textsc{Aya-13B}}} \\ \midrule[0.75pt]
      \textcolor{NavyBlue}{\textsc{Probs}} & .567 & .551 & .521 & .519 & .542 & .564 & .524 & .574 & .542 & \textbf{.626} & \underline{.597} & .567 & \underline{.555} & .547 & .513 & .560 & \textbf{.580} & \underline{.560} \\
      \textcolor{NavyBlue}{\textsc{Temp}} & .547 & .515 & .439 & .485 & .526 & .547 & .518 & .432 & .495 & .614 & \textbf{.610} & .556 & .543 & .489 & .559 & .527 & .556 & .549 \\
      \textcolor{NavyBlue}{\textsc{Ask Cali.}} & \underline{.613} & .503 & .494 & .476 & .474 & .492 & .490 & .460 & .486 & .512 & .510 & .489 & .508 & .466 & .496 & .514 & .490 & .496 \\
      \textcolor{DarkOrchid}{\textsc{Instruct}} & .539 & .441 & .348 & .412 & .362 & .417 & .426 & .419 & .404 & .559 & \underline{.597} & .421 & .510 & .333 & .481 & .442 & .480 & .466 \\
      \textcolor{Dandelion}{\textsc{Reflect}} & .410 & .347 & .300 & .339 & .336 & .357 & .335 & .347 & .337 & \underline{.615} & .489 & .357 & .448 & .312 & .437 & .404 & .426 & .410 \\
      \textcolor{Dandelion}{\textsc{MoreInfo}} & .409 & .348 & .299 & .350 & .357 & .358 & .337 & .368 & .345 & \underline{.615} & .497 & .360 & .444 & .325 & .441 & .393 & .413 & .410 \\
      \textcolor{Dandelion}{\textsc{BackTrans.}} & .450 & .421 & .333 & .453 & .346 & .354 & .411 & .411 & .390 & .542 & .571 & .393 & .484 & .300 & .487 & .442 & .474 & .450 \\
      \textcolor{OliveGreen}{\textsc{SCthres.}} & .609 & \underline{.618} & .614 & \underline{.609} & \underline{.610} & \textbf{.600} & .584 & .610 & \underline{.607} & .532 & .532 & .443 & \textbf{.577} & .543 & \underline{.572} & \textbf{.589} & .520 & .539 \\
      \textcolor{OliveGreen}{\textsc{Conflict}} & .564 & .567 & .581 & .568 & .521 & .568 & .561 & .582 & .564 & .536 & .520 & .546 & .514 & .559 & .548 & .553 & .544 & .540 \\ \midrule[0.75pt]
      \textcolor{Maroon}{\textsc{Mono-native}} & .512 & .580 & .515 & .604 & .529 & .576 & .533 & .520 & .551 & .552 & .578 & .479 & .452 & .467 & .481 & .524 & .526 & .501 \\
      \textcolor{Maroon}{\textsc{Mono-English}} & .611 & .611 & .607 & \textbf{.649} & .460 & .583 & .594 & \textbf{.688} & .599 & .581 & .513 & .514 & .503 & .513 & .506 & \underline{.565} & \underline{.572} & .527 \\
      \textcolor{Maroon}{\textsc{Multi-random}} & .540 & .597 & \underline{.615} & .561 & .524 & .549 & \underline{.628} & .605 & .583 & .481 & .403 & \textbf{.650} & .497 & \underline{.627} & .565 & \underline{.565} & .553 & .551 \\
      \textcolor{Maroon}{\textsc{Multi-related}} & \textbf{.631} & \textbf{.621} & \textbf{.704} & .595 & \textbf{.661} & \underline{.590} & \textbf{.643} & \underline{.628} & \textbf{.635} & .603 & .468 & \underline{.636} & .542 & \textbf{.693} & \textbf{.578} & .558 & .566 & \textbf{.577} \\ \midrule[0.75pt]
      \multicolumn{19}{c}{\textbf{\textsc{GPT-4}}} \\ \midrule[0.75pt]
      \textcolor{NavyBlue}{\textsc{Ask Cali.}} & .432 & .421 & .404 & .500 & .598 & .444 & .450 & .589 & .487 & .536 & .342 & .307 & .461 & .393 & .452 & .376 & .304 & .376 \\
      \textcolor{DarkOrchid}{\textsc{Instruct}} & \textbf{.789} & .566 & .363 & .493 & .386 & .556 & .481 & .465 & .473 & .656 & .552 & .186 & .432 & .160 & .435 & .272 & .270 & .330 \\
      \textcolor{Dandelion}{\textsc{Reflect}} & .686 & .655 & .585 & .649 & .528 & .597 & .519 & .589 & .589 & .658 & .545 & .229 & .561 & .347 & .571 & .483 & .408 & .449 \\
      \textcolor{Dandelion}{\textsc{MoreInfo}} & .694 & .572 & \textbf{.711} & .588 & \textbf{.677} & .611 & .558 & .612 & .619 & .386 & .461 & .486 & .555 & .507 & .584 & .469 & .543 & .515 \\
      \textcolor{Dandelion}{\textsc{BackTrans.}} & .764 & .634 & .563 & .696 & .535 & .660 & \underline{.620} & .636 & .621 & .538 & .522 & .576 & .564 & \textbf{.677} & .558 & \textbf{.555} & \textbf{.582} & \underline{.576} \\
      \textcolor{OliveGreen}{\textsc{SCthres.}} & .735 & .541 & .544 & .596 & .604 & .650 & .605 & .598 & .591 & \textbf{.759} & .508 & \textbf{.679} & .497 & \underline{.673} & .508 & .528 & .570 & .566 \\
      \textcolor{OliveGreen}{\textsc{Conflict}} & .730 & .555 & .641 & .589 & .561 & .629 & .559 & .590 & .589 & .639 & .488 & \underline{.593} & .503 & \underline{.673} & .501 & \underline{.535} & .557 & .550 \\ \midrule[0.75pt]
      \textcolor{Maroon}{\textsc{Mono-native}} & .728 & .655 & .548 & .642 & .567 & .660 & .589 & .628 & .613 & .708 & \underline{.558} & .371 & \underline{.665} & .307 & \underline{.597} & .401 & .447 & .478 \\
      \textcolor{Maroon}{\textsc{Mono-English}} & \textbf{.789} & .669 & .541 & \underline{.703} & .543 & .653 & .550 & \textbf{.659} & .617 & \underline{.737} & \textbf{.584} & .200 & .613 & .260 & .526 & .340 & .421 & .421 \\
      \textcolor{Maroon}{\textsc{Multi-random}} & .698 & \underline{.710} & .570 & .655 & .567 & \textbf{.681} & .581 & \underline{.651} & \underline{.631} & .714 & .532 & .300 & .606 & .380 & .532 & .408 & .441 & .457 \\
      \textcolor{Maroon}{\textsc{Multi-related}} & \underline{.785} & \textbf{.752} & \underline{.659} & \textbf{.730} & \underline{.638} & \underline{.674} & \textbf{.636} & \textbf{.659} & \textbf{.678} & .722 & .532 & .543 & \textbf{.706} & .647 & \textbf{.610} & .531 & \underline{.572} & \textbf{.592} \\ \midrule[0.75pt]
    \end{tabular}
    }
    \caption{Performance of \textcolor{NavyBlue}{calibration}, \textcolor{DarkOrchid}{training}, \textcolor{Dandelion}{prompting}, \textcolor{OliveGreen}{consistency}, and our proposed \textcolor{Maroon}{feedback}-based approaches on two LLMs and two multilingual datasets. We employ the Abstain Accuracy metric, Avg-H and Avg-L denote average performance for high and low-resource languages, while we additionally present performance for the seven low-resource languages (Bengali, Tamil, Nepali, Malayalam, Marathi, Telugu, and Kannada). Best performance in \textbf{bold} and second-best in \underline{underline}. Baselines that rely on token probabilities (e.g., \textcolor{NavyBlue}{Probs}) are not compatible with GPT-4. \textbf{\textsc{Multi-related} achieves the best average performance in low-resource languages across all models and datasets, improving over baselines by up to 9.2\%.}}
    %\vspace*{-10pt}
    \label{tab:big}
\end{table*}

\section{Experiment Settings}

\paragraph{Models} We evaluate existing approaches and the four proposed monolingual/multilingual feedback strategies with three LLMs: \emph{Aya-13B}, a specifically multilingual instruction-tuned model, \emph{ChatGPT} and \emph{GPT-4}, two general-purpose black-box LLMs. We employ greedy decoding for QA and making an abstain decision, and employ a temperature of 0.7 when sampling repeatedly (e.g., consistency-based baselines and feedback generation).

\paragraph{Datasets} We evaluate with the Multilingual MMLU (M-MMLU) and Hellaswag (M-Hellaswag) datasets \citep{lai2023okapi}, featuring encyclopedic and commonsense knowledge. Originally in English, these QA problems were translated into 26 other languages through machine translation. These languages are characterized as 8 high-resource languages, 11 mid-resource languages, and 7 low-resource languages based on their proportion in pretraining data.\footnote{Full language list in Appendix \ref{appendix:experiment_details}.} We also present evaluation with Belebele \citep{bandarkar2023belebele} in Appendix \ref{appendix:analysis_cont}, a multilingual reading comprehension dataset. For the three datasets, we create random splits with 200 instances for validation and 800 for test, with minor variation across languages due to data availability.

\paragraph{Baselines} We compare with nine abstain baselines that could be adapted in multilingual settings: calibration-based \textsc{Probs} (token probabilities), \textsc{Temp} \citep{jiang2021can}, \textsc{Ask Cali.} \citep{tian2023just}; training-based \textsc{Instruct} \citep{ouyang2022training}; prompting-based \textsc{Reflect} \citep{kadavath2022language}, \textsc{MoreInfo} \citep{Feng2023KnowledgeCF}, \textsc{BackTrans} \citep{edunov2018understanding}; and consistency-based approaches \textsc{SCthres.} \citep{wang2022self}, \textsc{Conflict} \citep{feng2024don}. More details about the baselines are in Appendix \ref{appendix:experiment_details}.

\paragraph{Evaluation Metrics} We use the Abstain Accuracy metric (A-Acc) proposed in \citet{feng2024don}: LLMs should abstain when it would provide an incorrect answer and should not abstain when it would provide a correct answer, concretely $\textit{A-Acc} = \frac{TP+TN}{TP+TN+FP+FN}$ and $TP$ indicates the LLM should abstain and did. We additionally report other AbstainQA metrics (Reliable Accuracy, Effective Reliability) in Appendix \ref{appendix:analysis_cont}.

\section{Results}
\label{sec:results}

We present the abstain accuracy results with two LLMs on two multilingual datasets in Table \ref{tab:big}.

\paragraph{\textsc{Multi-related} achieves state-of-the-art performance.} \textsc{Multi-related} achieves the highest average performance on low-resource languages (Avg-L) across all four model and dataset settings, improving over the strongest baseline by 4.9\% on average. Out of the 7 low-resource languages, \textsc{multi-related} achieves the best and top-2 performance in 3.25 and 4.75 languages on average. This improvement in low-resource languages comes with on-par performance in high-resource languages (Avg-H), outperforming baselines in 81\% of the times across four (model, dataset) settings. This indicates that by generating and reflecting on multilingual feedback from related languages, LLMs greatly improve in identifying inherent knowledge gaps across languages.

\paragraph{Existing approaches greatly drop beyond high-resource languages.} Ask for Calibration \citep{tian2023just}, an approach to solicit LLM confidence scores verbally, witness a 12.7\% drop from high to low-resource languages (0.613 $\rightarrow$ 0.486) on MMLU using \textsc{Aya-13B}. While it could generate meaningful confidence scores between 0 and 1 for high-resource languages, it collapses and repeatedly generate the same number (e.g., 0.8) for almost all questions in low-resource languages. Similar performance gaps and failure modes could be observed for previously strong approaches in English such as Instruction Tuning (35.3\% drop, on average), Self-Reflect (33.3\%), and SCthreshold (12.2\%). In comparison, \textsc{Mult-related} has a smaller drop of 8.5\%: we further quantify the fairness of abstain strategies in Section \ref{sec:analysis}.

\paragraph{Abstaining is a language-specific problem.} Out of the seven low-resource languages, we observe that Tamil (ta) and Malayalam (ml) are consistently the most challenging languages across models, datasets, and approaches: an average performance of 0.484 and 0.492 is achieved on the two languages, while the global average for low-resource languages is 0.520. This could be attributed to their low representation in LLM pretraining data \citep{lai2023okapi} and thus lower utility, meaning that there is no one-size-fits-all solution for abstaining across multilingual contexts and robust strategies should be language-specific. \textsc{Multi-related} takes linguistic knowledge into account by employing \emph{related} languages for feedback generation, successfully achieving the best Avg-L performance across all models and datasets. We further study the utility of language relatedness in Section \ref{sec:analysis}.

\paragraph{\textsc{Aya-13B} shows smaller gaps than \textsc{GPT-4}.} While the performance of \textsc{Multi-related} is higher on GPT-4, the gap between low and high-resource languages is smaller with \textsc{Aya-13B} (1.7\% vs.~16.9\%). Since \textsc{Multi-related} specifically relies on generating and reasoning in multilingual contexts, the explicitly multilingual \textsc{Aya-13B} would be better than the general-purpose \textsc{GPT-4} to this end. This motivates a potential collaboration between models: using a stronger general-purpose LLM for QA and a smaller but explicitly multilingual LLM for feedback generation. We further explore this in Section \ref{sec:analysis}.

\section{Analysis}
\label{sec:analysis}

\paragraph{\textsc{Multi-related} is more equitable.} While we primarily focused on the performance gaps between high and low-resource languages in Section \ref{sec:results}, measuring the fairness of a multilingual system goes beyond performance averages. Concretely, we follow \citet{song2023globalbench} to measure utility and equity, indicating how well multilingual approaches serve diverse language speakers and performance disparity across languages. For utility:
\begin{align*}
    \mathrm{M}_{\tau} = \sum_{\boldsymbol{\ell} \in \mathcal{L}}        {\mathrm{d}_{\boldsymbol{\ell}}}^\tau \cdot \mathrm{u}_{\boldsymbol{\ell}}, \ \ \ \mathrm{d}_{\boldsymbol{\ell}} = \frac{\mathrm{n}_{\boldsymbol{\ell}}}{\sum_{\boldsymbol{\ell} \in \mathcal{L}} \mathrm{n}_{\boldsymbol{\ell}}}
\end{align*}
where $\mathrm{u}_{\boldsymbol{\ell}}$ denotes the utility/performance on language $\boldsymbol{\ell}$, $\mathrm{n}_{\boldsymbol{\ell}}$ denotes the number of native speakers, the exponential $\tau = 1$ indicates \emph{demographic weighted} utility and $\tau = 0$ indicates \emph{lingustic weighted} utility where all languages are treated as equals. For equity, performance on various languages are sorted in non-decreasing order ($\mathrm{u}_i \leq \mathrm{u}_{i+1}$) and the Gini coefficient is calculated:

\begin{align*}
    \mathrm{G} = \frac{1}{\mid \mathcal{L} \mid} \Bigl ( \mid \mathcal{L} \mid + 1 - 2\frac{\sum_{i=1}^{\mid \mathcal{L} \mid} (\mid \mathcal{L} \mid + 1 - i)\mathrm{u}_i}{\sum_{i=1}^{\mid \mathcal{L} \mid}\mathrm{u}_i} \Bigr )
\end{align*}
where $\mid \mathcal{L} \mid$ indicates the total number of languages. The range of $\mathrm{G}$ is 0 to 1 and more equitable abstain strategies should have lower $\mathrm{G}$ values.

We present the demographic utility, linguistic utility, and equity metrics in Table \ref{tab:fairness}. \textsc{Multi-related} outperforms baselines on both utility modes, while being more equitable across languages, evident in the 12.9\% reduction in Gini Coefficient. On the contrary, \textsc{Mono-English} have on-par demographic utility but worse linguistic utility and equity, indicating that generated feedback in English is unevenly helpful to other languages, whereas low-resource languages distant from English benefit much less.

\begin{table}[t]
    \centering
    \setlength{\tabcolsep}{3pt}
    \renewcommand{\arraystretch}{0.9}
    \resizebox{1\linewidth}{!}{
    \begin{tabular}{lccc}
         \toprule[1.5pt]
         \textbf{Method} &\textbf{Demo.} ($\mathrm{M}_1, \uparrow$) &\textbf{Ling.} ($\mathrm{M}_0, \uparrow$) &\textbf{Equity} ($\mathrm{G}, \downarrow$) \\ \midrule[0.75pt]
            \textcolor{NavyBlue}{\textsc{Probs}} &0.5613 &0.5632 &\underline{0.0319} \\
            \textcolor{NavyBlue}{\textsc{Ask Cali.}} &0.5976 &0.5784 &0.0488 \\
            \textcolor{DarkOrchid}{\textsc{Instruct}} &0.4514 &0.4280 &0.0477 \\
            \textcolor{Dandelion}{\textsc{Reflect}} &0.3983 &0.3877 &0.0460 \\
            \textcolor{Dandelion}{\textsc{Backtrans.}} &0.4342 &0.4261 &0.0517 \\
            \textcolor{OliveGreen}{\textsc{SCThres.}} &0.5974 &\underline{0.5916} &0.0340 \\
            \textcolor{OliveGreen}{\textsc{Conflict}} &0.5698 &0.5630 &0.0369 \\ \midrule[0.75pt]
            \textcolor{Maroon}{\textsc{Mono-native}} &0.5181 &0.5318 &0.0472 \\
            \textcolor{Maroon}{\textsc{Mono-English}} &\underline{0.6038} &0.5651 &0.0564 \\
            \textcolor{Maroon}{\textsc{Multi-random}} &0.5442 &0.5528 &0.0390 \\
            \textcolor{Maroon}{\textsc{Multi-related}} &\textbf{0.6149} &\textbf{0.6027} &\textbf{0.0278} \\\bottomrule[1.5pt]
    \end{tabular}
    }
    \caption{Utility and equity metrics of abstain strategies, where $\uparrow$/$\downarrow$ indicates that higher/lower values are desirable. Best performance in \textbf{bold} and second-best in \underline{underline}. \textbf{\textsc{Multi-related} offers a fairer abstain strategy with higher utility and lower Gini coefficient.}}
    %\vspace*{-10pt}
    \label{tab:fairness}
\end{table}

\begin{figure}
    \centering
    \includegraphics[width=1\linewidth]{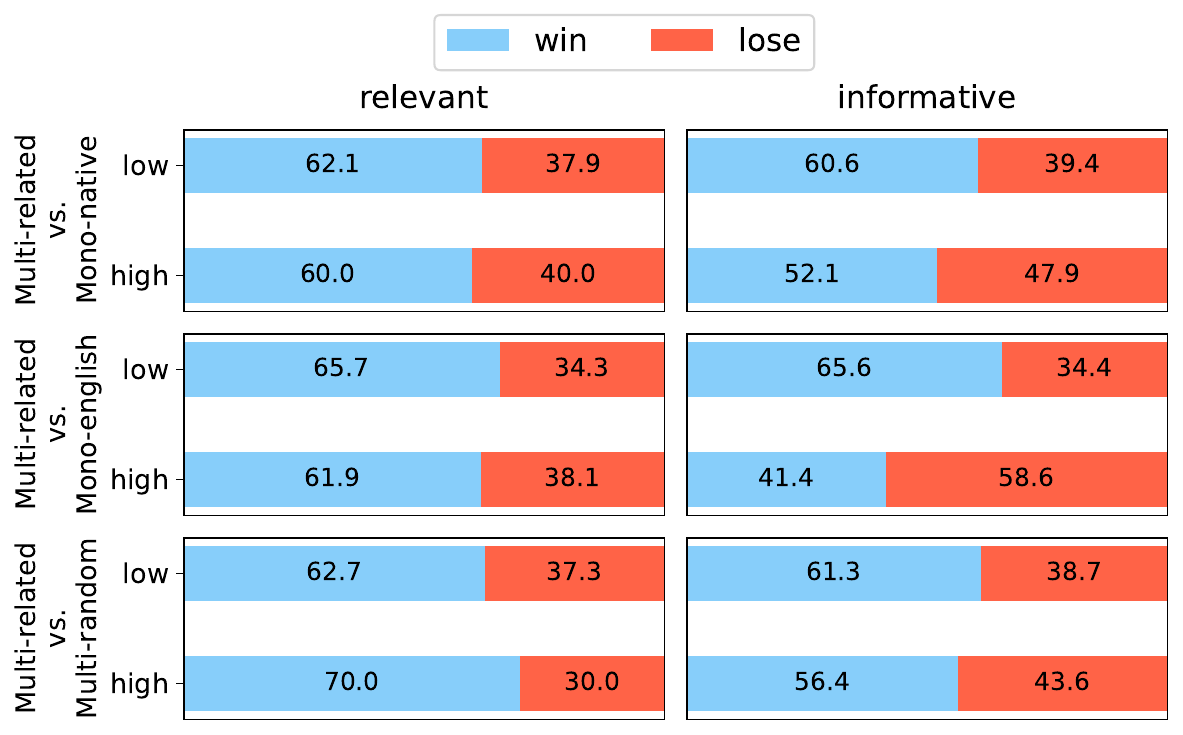}
    \caption{GPT-4 evaluation of feedback pairs to the same question, comparing \textsc{Multi-related} against other feedback settings to evaluate which produces more \emph{relevant} and \emph{informative} feedback.}
    %\vspace*{-10pt}
    \label{fig:quality}
\end{figure}

\paragraph{\textsc{Multi-related} offers relevant, informative, and conflicting pieces of feedback.} To better understand the quality and role of the generated feedback, we employ GPT-4 evaluation for large-scale automatic analysis. We compare the four feedback modes by first translating all feedback into English with GPT-4 to remove the confounding factor of language difference, then using LLM-as-a-judge methodology \citep{zheng2024judging} for pairwise comparison. Given a question, proposed answer, and a pair of two feedback from \textsc{Multi-related} and a baseline, GPT-4 evaluates which feedback is more \emph{relevant} and \emph{informative}. We present the win rate in Figure \ref{fig:quality}: \textsc{Multi-related} is consistently more relevant to the question across languages. While \textsc{Mono-English} offers more informative feedback on high-resource languages, it degrades into the worst on low-resource languages while \textsc{Multi-related} becomes the best.

\begin{figure}
    \centering
    \includegraphics[width=1\linewidth]{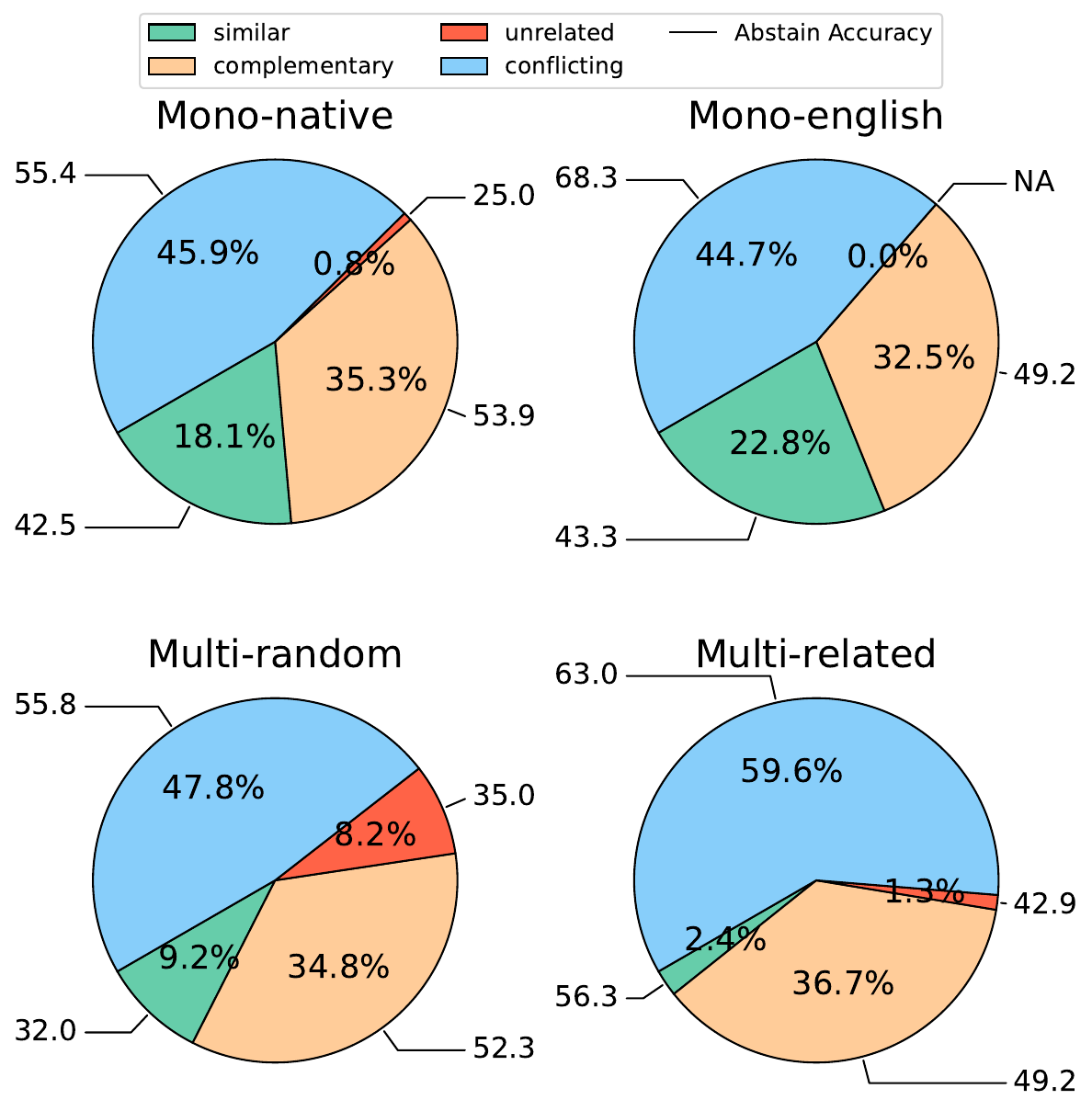}
    \caption{\textbf{GPT-4 evaluation of the role of the $k$ pieces of feedback to a given question}: whether they are similar, complementary, conflicting, or unrelated. We also present the abstain accuracy in each feedback role below the role name, showing that conflicting feedback are best for LLM self-reflection while \emph{Multi-related} has the most conflicting feedback scenarios.}
    %\vspace*{-10pt}
    \label{fig:role}
\end{figure}

For feedback roles, we first conduct a manual examination to identify four potential roles of the $k$ generated feedback: \emph{similar}, \emph{complementary}, \emph{conflicting}, and \emph{unrelated}. We then employ GPT-4 to evaluate the roles of $k$ feedback, translated into English, for the same question. Figure \ref{fig:role} demonstrates that monolingual approaches result in 252.7\% more similar and thus redundant feedback compared to multilingual settings, while \textsc{Multi-random} result in greater unrelated feedback potentially due to the random selection of distant languages. In comparison, \textsc{Multi-related} produces 24.7\% more conflicting scenarios where feedback disagree in content or conclusion: the abstain accuracy on \emph{conflicting} scenarios are also the highest, indicating that LLMs face more knowledge conflicts \citep{xie2023adaptive, wang2023resolving} by generating multiple feedback from related but different languages, which in turn aids self-reflection and making better-informed abstain decisions \citep{feng2024don}. We further present a qualitative analysis in Appendix \ref{appendix:analysis_cont} in addition to the automatic GPT-4 evaluation.

\paragraph{Culture is a driving factor in multilingual abstention.} For \textsc{Multi-related}, we by default define language relatedness as the average of the six linguistic attributes in Lang2vec \citep{littell2017uriel}. (\Sref{sec:methodology}) We further investigate what aspects of ``language relatedness'' are most helpful for abstaining across multilingual contexts. Specifically, we additionally select related languages only by one of the six categories (e.g., syntactic or phonological relatedness). We introduce two additional settings: 1) LLMs are prompted to propose three related languages by themselves; 2) related languages in the same \emph{culture} cluster according to the World Value Survey.\footnote{\href{https://www.worldvaluessurvey.org}{https://www.worldvaluessurvey.org}} We present the performance of various language relatedness settings in Table \ref{tab:relatedness_study}. We observe that geography and phonology are the most helpful linguistic attributes, while \emph{culture}-informed language selection yields the best utility and equity results. This indicates that \emph{multilingual feedback from languages of related socio-cultural backgrounds is most helpful for low-resource languages and overall fairness.}

\begin{table}[t]
    \centering
    \resizebox{1\linewidth}{!}{
    \begin{tabular}{lcccc}
         \toprule[1.5pt]
         \textbf{Setting} & \textbf{Avg-H} & \textbf{Avg-M} & \textbf{Avg-L} & \textbf{Equity} ($\downarrow$) \\ \midrule[0.75pt]
         \textsc{Default} &0.6411 &0.5861 &0.4432 &0.0943 \\
         \textsc{Syntactic} &0.6452 &0.5843 &0.4395 &0.0967 \\
         \textsc{Geographic} &\textbf{0.6589} &0.5892 &0.4712 &0.0880 \\
         \textsc{Phonological} &\underline{0.6499} &0.5881 &\underline{0.5017} &\underline{0.0694} \\
         \textsc{Genetic} &0.6476 &0.6024 &0.4944 &0.0839 \\
         \textsc{Inventory} &0.6371 &0.5827 &0.4356 &0.0950 \\
         \textsc{Featural} &0.6412 &\underline{0.6116} &0.4417 &0.0916 \\
         \textsc{LLM-generated} &0.6316 &0.5929 &0.4362 &0.0981 \\
         \textsc{Culture} &0.6425 &\textbf{0.6202} &\textbf{0.5322} &\textbf{0.0438} \\\bottomrule[1.5pt]
    \end{tabular}
    }
    \caption{Performance averages for high, mid, and low-resource languages, as well as the equity metric $\mathrm{G}$ for various language relatedness settings. Best performance in \textbf{bold} and second-best in \underline{underline}. \textbf{Culturally informed language selection is best for mid and low-resource languages and also more equitable.}}
    %\vspace*{-10pt}
    \label{tab:relatedness_study}
\end{table}

\begin{figure*}
    \centering
    \includegraphics[width=0.9\linewidth]{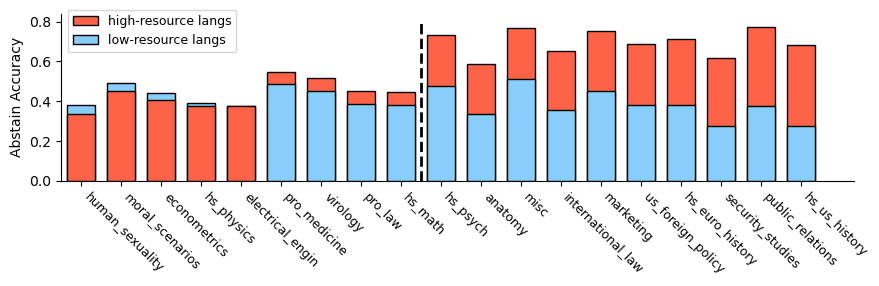}
    \caption{Abstain accuracy on various MMLU domains with high and low-resource languages: on the left we show the 10 domains with the least performance gaps and on the right we show the 10 with the most gaps. ``hs'' indicates high school. \textbf{While domains with the least gaps often feature STEM topics that are more objective, domains with the largest gaps are often driven by culture, especially West-centric social knowledge.}}
    %\vspace*{-10pt}
    \label{fig:mmlu_domains}
\end{figure*}

To further investigate the impact of culture, we present the performance breakdown of various MMLU domains in Figure \ref{fig:mmlu_domains}. We illustrate the 10 domains with the largest gaps between low- and high-resource languages and 10 domains with the least gaps. The largest gaps often come from west-centric topics such as ``US history'', ``European history'', and ``US foreign policy'', while the smallest gaps are often on STEM domains that transcend socio-cultural contexts such as ``logical fallacies'', ``high school physics'', and ``electrical engineering''. This again indicates that \emph{culture} is a driving factor in multilingual abstention: improving LLM abstain capabilities is not only a technical problem but also a social-oriented one, where the existing West-centric LLMs \citep{naous2023having} should better incorporate other cultures and perspectives for equitable improvements in factuality and reliability.

\begin{figure}
    \centering
    \includegraphics[width=1\linewidth]{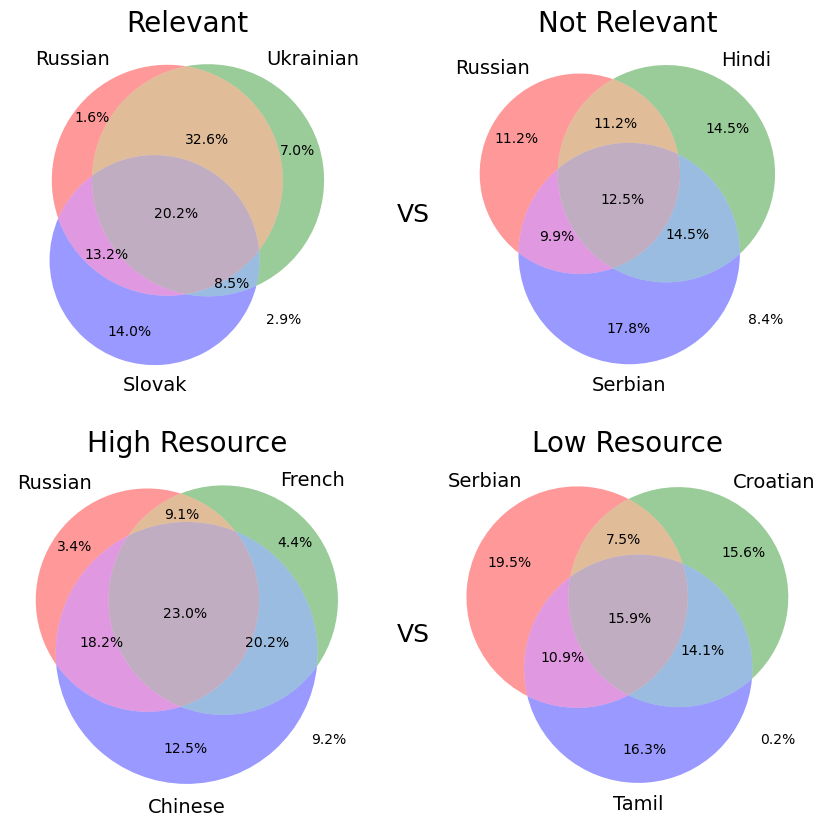}
    \caption{Overlap of abstain decisions made in different languages, where the overlap indicates that LLMs abstain in both/all three of the languages. \textbf{We find that abstain decisions are only somewhat transferrable between relevant and high-resource language clusters.}}
    %\vspace*{-10pt}
    \label{fig:overlap}
\end{figure}

\paragraph{Abstain decisions are less transferable across unrelated and low-resource languages.} One solution to multilingual abstain is to take the highest-resource language (e.g., English), make abstain decisions, and use that decision to abstain/generate in low-resource languages. However, to what extent do abstain decisions overlap across languages and thus transferable remains underexplored, which could not be taken for granted given the factuality variation across languages \citep{lai2023okapi, kang2024comparing}. To this end, we visualize the abstain overlap of parallel questions across various three-language groups in Figure \ref{fig:overlap}, where overlapping parts indicate that $\textsc{Multi-related}$ for 2 or 3 languages decided to abstain. For control group \#1, the group of three related languages sees much greater overlap (74.5\% 2+ overlap) than the three unrelated languages (48.1\%). For control group \#2, a group of three high-resource languages sees greater overlap (70.5\%) than three low-resource languages (48.4\%). These two findings together indicate that abstain decisions are only somewhat transferable in the case of high-resource closely related languages: however, many languages on the long tail are neither close to English nor well-represented in LLM training data, thus English-only abstain methods are not one-size-fit-all solutions and abstaining is a language-specific problem.

\begin{table}[t]
    \centering
    \resizebox{1\linewidth}{!}{
    \begin{tabular}{l|ccccccccc}
         \toprule[1.5pt]
         \multirow{2}{*}{\textbf{Method}} & \multicolumn{9}{c}{\textbf{High-Resource}} \\\cmidrule{2-10}
         &ru &de &zh &fr &es &it &nl &vi &avg. \\\midrule[0.75pt]
         \textsc{Self} &.818 &.852 &.794 &.838 &.823 &.769 &.831 &.555 &\bf .785 \\
         \textsc{Other} &.733 &.788 &.762 &.772 &.815 &.733 &.808 &.668 &.760 \\\midrule[0.75pt]
         \multirow{2}{*}{\textbf{Method}} & \multicolumn{9}{c}{\textbf{Low-Resource}} \\\cmidrule{2-10}
         &bn &ta &ne &ml &mr &te &kn &/ &avg. \\\midrule[0.75pt]
         \textsc{Self} &.752 &.659 &.730 &.638 &.674 &.636 &.659 &/ &.678 \\
         \textsc{Other} &.788 &.722 &.735 &.656 &.669 &.735 &.697 &/ &\bf .715 \\
         \bottomrule[1.5pt]
    \end{tabular}
    }
    \caption{Performance when using \textsc{GPT-4} it\emph{self} or the \emph{other} \textsc{Aya-13B} multilingual LLM for feedback generation. The collaboration between a general-purpose LLM for QA and a smaller but more multilingual model for feedback generation benefits low-resource languages.}
    %\vspace*{-10pt}
    \label{tab:other_llm}
\end{table}

\paragraph{General-purpose LLMs could be supervised by a smaller but more multilingual model.} Motivated by the finding that \emph{GPT-4} has higher absolute performance but \emph{Aya-13B} witnesses smaller gaps with $\textsc{Multi-related}$ (\Sref{sec:results}), we explore the collaboration between the two models: using \emph{GPT-4} for question answering and \emph{Aya-13B} for multilingual feedback generation. Table \ref{tab:other_llm} demonstrates that while for high-resource languages this might be counterproductive, for low-resource languages it results in a 5.4\% improvement on average. This indicates that when user queries come in low-resource and underrepresented languages, a smaller but explicitly multilingual model could be employed to supervise general-purpose black-box LLMs to improve abstaining and mitigate hallucinations.

\section{Related Work}

\paragraph{Teaching LLMs to Abstain} Existing works focus on various types of approaches: \emph{Calibration-based} approaches focus on eliciting the confidence levels of LLMs with token probabilities \citep{sun2022quantifying, zhou2023batch, liu2023litcab} or semantic markers of uncertainty \citep{kuhn2023semantic, zhou-etal-2023-navigating, Zhou2024RelyingOT}, where previous research evaluate \citep{radford2019language, ahuja-etal-2022-calibration, liang2022holistic, tao2023benchmark, he2023investigating} and improve \citep{kamath-etal-2020-selective, desai-durrett-2020-calibration, jagannatha-yu-2020-calibrating, kong-etal-2020-calibrated, jiang2021can, lin2022teaching, mielke2022reducing} calibration for various tasks \citep{wang-etal-2020-inference, stengel2023calibrated, kalai2023calibrated, zablotskaia-etal-2023-uncertainty}. \emph{Prompting-based} approaches induce self-reflection by including none-of-the-above options \citep{kadavath2022language}, prompt to self-correct reasoning \citep{kim2023language, shinn2023reflexion, huang2023large, chen-etal-2023-adaptation}, ask for additional information \citep{Feng2023KnowledgeCF}, and more \citep{wang-etal-2023-chatgpt-defend, si2023getting}. \emph{Training-based} approaches aim to adapt LLMs for abstention with linear probing \citep{slobodkin2023curious, azaria2023internal}, training an extra module \citep{cobbe2021training}, or alignment objectives \citep{zhang2023r, yang2023alignment, sun2023aligning, bashlovkina2023trusted, huang2023survey, liu2023trustworthy}. \emph{Consistency-based} \citep{wang2022self, cole-etal-2023-selectively} and \emph{collaboration-based} approaches \citep{feng2024don} are also explored to gauge LLM confidence through output variation from a single model or knowledge variation across multiple models. However, most existing approaches were proposed and evaluated with English only, while Figure \ref{fig:teaser} shows that multilingual abstention poses new challenges to existing solutions and leads to performance gaps based on language resourcesness. To mitigate this gap, we propose to teach LLMs to abstain by learning from multilingual feedback, where diverse feedback are generated in related languages to enhance reliability and make trustworthy abstain decisions.

\vspace*{-5pt}
\paragraph{Multilingual Factuality} While early factuality studies were mostly conducted in English \citep{huang2023survey, zhang2023siren, ji2023survey}, understanding and mitigating hallucinations beyond English is increasingly important for LLMs to equitably serve diverse populations \citep{liu-etal-2022-enhancing-multilingual, lai2023okapi, xu-etal-2023-condensing, xu-etal-2023-language-representation, qi-etal-2023-cross, schott-etal-2023-polyglot, kang2024comparing, gao-etal-2024-multilingual}. In multilingual summarization, metrics and evaluations are proposed to quantify factual errors and utility \citep{aharoni2022mface, qiu2023detecting, clark-etal-2023-seahorse}. In machine translation, faithfulness across diverse languages is also a critical concern \citep{lee2018hallucinations, raunak2021curious, xu2023understanding, dale2023detecting, dale-etal-2023-halomi}. A diverse range of models \citep{lin-etal-2022-shot, muennighoff2023crosslingual, lai2023okapi, ustun2024aya}, datasets \citep{artetxe2020cross, clark2020tydi, longpre2021mkqa, chalkidis-etal-2022-fairlex, gehrmann-etal-2022-gemv2, ebrahimi-etal-2022-americasnli, li-etal-2022-multi-level, asai2023buffet, ogundepo2023cross, ahuja-etal-2023-mega, wang-etal-2024-seaeval}, and studies on multilingual transfer \cite{lin-etal-2019-choosing, pires-etal-2019-multilingual, wu-dredze-2019-beto, karthikeyan2019cross, wu-etal-2022-learning, fujinuma-etal-2022-match, ustun-etal-2022-hyper, schmidt-etal-2022-dont, asai2023buffet, philippy-etal-2023-towards, tanwar-etal-2023-multilingual, reusens-etal-2023-investigating, li-etal-2024-improving-context, gao-etal-2024-multilingual} also contribute to the improvement of LLM factuality and utility beyond English. In this work, we present the first study on LLM abstaining in multilingual contexts and make an important step toward improving the reliability of multilingual LLMs and mitigating hallucinations.

\section{Conclusion}
We propose to improve the reliability of multilingual LLMs by abstaining via \emph{multilingual feedback}, where LLMs generate feedback to their proposed answer in related languages for self-reflection. Extensive experiments demonstrate that \emph{multilingual feedback} achieves up to 9.2\% improvement against baselines across models and datasets, while presenting a more equitable solution to multilingual abstention. Further analysis reveals that abstention is a language-specific problem, that multilingual feedback in related languages both improves the accuracy of abstention and calibrates the fairness across higher- and lower-resource languages, and that cultural relatedness is an important factor in the utility and equity of abstention, highlighting that multilingual modeling is not only a technical problem but also a social-oriented one.

\section*{Limitations}
Our study of teaching LLMs to abstain focuses on the \emph{knowledge perspective}, i.e., LLMs should abstain when their parametric knowledge is insufficient to provide a correct answer. However, the abstain problem also has implications from the \emph{safety} perspective \citep{huang2023survey, liu2023trustworthy}. We envision future methodologies and evaluations that tackle both directions of the abstain problem across diverse language contexts.

Our approach, teaching LLMs to abstain via multilingual feedback, involves sampling multiple feedback from related languages to promote self-reflection. This sampling introduces minor randomness in LLMs' abstain decisions (Appendix \ref{appendix:analysis_cont}). In addition, it would incur greater inference costs than the most simple prompting approaches, but is also not the most expensive \citep{feng2024don}. When a black-box LLM with hundreds of billions of parameters is served behind an API call, our approach enables the incorporation of one extra multilingual 7B model for stronger reliability (Table \ref{tab:other_llm}) and does not add much to the overall cost.

\section*{Ethics Statement}
While abstaining in multilingual contexts is a technical problem, we discover the role of culture in AbstainQA and that west-centric LLMs \citep{naous2023having, li2024culturepark, rao2024normad} are hindering progress on equitable LLM abstention (\Sref{sec:analysis}). This encourages research at the intersection of multilingualism and culture \citep{choenni2024echoes}. We envision future work on not only proposing technical solutions to the abstain problem, but also improving the representation of diverse values, perspectives, and cultures in LLMs.

\section*{Acknowledgements}
We gratefully acknowledge support from the National Science Foundation under CAREER Grant No.~IIS2142739, and NSF grants No.~IIS2125201 and IIS2203097. This work was also supported in part by gift funding from Google, MSR, and Microsoft Accelerate Foundation Models Research.

\bibliography{custom}

\begin{thebibliography}{127}
\providecommand{\natexlab}[1]{#1}

\bibitem[{Aharoni et~al.(2022)Aharoni, Narayan, Maynez, Herzig, Clark, and Lapata}]{aharoni2022mface}
Roee Aharoni, Shashi Narayan, Joshua Maynez, Jonathan Herzig, Elizabeth Clark, and Mirella Lapata. 2022.
\newblock mface: Multilingual summarization with factual consistency evaluation.
\newblock \emph{arXiv preprint arXiv:2212.10622}.

\bibitem[{Ahuja et~al.(2023)Ahuja, Diddee, Hada, Ochieng, Ramesh, Jain, Nambi, Ganu, Segal, Ahmed, Bali, and Sitaram}]{ahuja-etal-2023-mega}
Kabir Ahuja, Harshita Diddee, Rishav Hada, Millicent Ochieng, Krithika Ramesh, Prachi Jain, Akshay Nambi, Tanuja Ganu, Sameer Segal, Mohamed Ahmed, Kalika Bali, and Sunayana Sitaram. 2023.
\newblock {MEGA}: Multilingual evaluation of generative {AI}.
\newblock In \emph{Proceedings of the 2023 Conference on Empirical Methods in Natural Language Processing}.

\bibitem[{Ahuja et~al.(2022)Ahuja, Sitaram, Dandapat, and Choudhury}]{ahuja-etal-2022-calibration}
Kabir Ahuja, Sunayana Sitaram, Sandipan Dandapat, and Monojit Choudhury. 2022.
\newblock On the calibration of massively multilingual language models.
\newblock In \emph{Proceedings of the 2022 Conference on Empirical Methods in Natural Language Processing}.

\bibitem[{Artetxe et~al.(2020)Artetxe, Ruder, and Yogatama}]{artetxe2020cross}
Mikel Artetxe, Sebastian Ruder, and Dani Yogatama. 2020.
\newblock On the cross-lingual transferability of monolingual representations.
\newblock In \emph{Proceedings of the 58th Annual Meeting of the Association for Computational Linguistics}, pages 4623--4637.

\bibitem[{Asai et~al.(2023)Asai, Kudugunta, Yu, Blevins, Gonen, Reid, Tsvetkov, Ruder, and Hajishirzi}]{asai2023buffet}
Akari Asai, Sneha Kudugunta, Xinyan~Velocity Yu, Terra Blevins, Hila Gonen, Machel Reid, Yulia Tsvetkov, Sebastian Ruder, and Hannaneh Hajishirzi. 2023.
\newblock Buffet: Benchmarking large language models for few-shot cross-lingual transfer.
\newblock \emph{arXiv preprint arXiv:2305.14857}.

\bibitem[{Asai et~al.(2021)Asai, Yu, Kasai, and Hajishirzi}]{asai2021one}
Akari Asai, Xinyan Yu, Jungo Kasai, and Hanna Hajishirzi. 2021.
\newblock One question answering model for many languages with cross-lingual dense passage retrieval.
\newblock \emph{Advances in Neural Information Processing Systems}, 34:7547--7560.

\bibitem[{Azaria and Mitchell(2023)}]{azaria2023internal}
Amos Azaria and Tom Mitchell. 2023.
\newblock The internal state of an llm knows when it’s lying.
\newblock In \emph{Findings of the Association for Computational Linguistics: EMNLP 2023}, pages 967--976.

\bibitem[{Bandarkar et~al.(2023)Bandarkar, Liang, Muller, Artetxe, Shukla, Husa, Goyal, Krishnan, Zettlemoyer, and Khabsa}]{bandarkar2023belebele}
Lucas Bandarkar, Davis Liang, Benjamin Muller, Mikel Artetxe, Satya~Narayan Shukla, Donald Husa, Naman Goyal, Abhinandan Krishnan, Luke Zettlemoyer, and Madian Khabsa. 2023.
\newblock The belebele benchmark: a parallel reading comprehension dataset in 122 language variants.
\newblock \emph{arXiv preprint arXiv:2308.16884}.

\bibitem[{Bashlovkina et~al.(2023)Bashlovkina, Kuang, Matthews, Clifford, Jun, Cohen, and Baumgartner}]{bashlovkina2023trusted}
Vasilisa Bashlovkina, Zhaobin Kuang, Riley Matthews, Edward Clifford, Yennie Jun, William~W Cohen, and Simon Baumgartner. 2023.
\newblock Trusted source alignment in large language models.
\newblock \emph{arXiv preprint arXiv:2311.06697}.

\bibitem[{Brown et~al.(2020)Brown, Mann, Ryder, Subbiah, Kaplan, Dhariwal, Neelakantan, Shyam, Sastry, Askell, Agarwal, Herbert-Voss, Krueger, Henighan, Child, Ramesh, Ziegler, Wu, Winter, Hesse, Chen, Sigler, Litwin, Gray, Chess, Clark, Berner, McCandlish, Radford, Sutskever, and Amodei}]{brown2020language}
Tom Brown, Benjamin Mann, Nick Ryder, Melanie Subbiah, Jared~D Kaplan, Prafulla Dhariwal, Arvind Neelakantan, Pranav Shyam, Girish Sastry, Amanda Askell, Sandhini Agarwal, Ariel Herbert-Voss, Gretchen Krueger, Tom Henighan, Rewon Child, Aditya Ramesh, Daniel Ziegler, Jeffrey Wu, Clemens Winter, Chris Hesse, Mark Chen, Eric Sigler, Mateusz Litwin, Scott Gray, Benjamin Chess, Jack Clark, Christopher Berner, Sam McCandlish, Alec Radford, Ilya Sutskever, and Dario Amodei. 2020.
\newblock Language models are few-shot learners.
\newblock In \emph{Advances in Neural Information Processing Systems}.

\bibitem[{Chalkidis et~al.(2022)Chalkidis, Pasini, Zhang, Tomada, Schwemer, and S{\o}gaard}]{chalkidis-etal-2022-fairlex}
Ilias Chalkidis, Tommaso Pasini, Sheng Zhang, Letizia Tomada, Sebastian Schwemer, and Anders S{\o}gaard. 2022.
\newblock {F}air{L}ex: A multilingual benchmark for evaluating fairness in legal text processing.
\newblock In \emph{Proceedings of the 60th Annual Meeting of the Association for Computational Linguistics (Volume 1: Long Papers)}.

\bibitem[{Chen et~al.(2023)Chen, Yoon, Ebrahimi, Arik, Pfister, and Jha}]{chen-etal-2023-adaptation}
Jiefeng Chen, Jinsung Yoon, Sayna Ebrahimi, Sercan Arik, Tomas Pfister, and Somesh Jha. 2023.
\newblock Adaptation with self-evaluation to improve selective prediction in {LLM}s.
\newblock In \emph{Findings of the Association for Computational Linguistics: EMNLP 2023}.

\bibitem[{Choenni et~al.(2024)Choenni, Lauscher, and Shutova}]{choenni2024echoes}
Rochelle Choenni, Anne Lauscher, and Ekaterina Shutova. 2024.
\newblock The echoes of multilinguality: Tracing cultural value shifts during lm fine-tuning.
\newblock \emph{arXiv preprint arXiv:2405.12744}.

\bibitem[{Clark et~al.(2023)Clark, Rijhwani, Gehrmann, Maynez, Aharoni, Nikolaev, Sellam, Siddhant, Das, and Parikh}]{clark-etal-2023-seahorse}
Elizabeth Clark, Shruti Rijhwani, Sebastian Gehrmann, Joshua Maynez, Roee Aharoni, Vitaly Nikolaev, Thibault Sellam, Aditya Siddhant, Dipanjan Das, and Ankur Parikh. 2023.
\newblock {SEAHORSE}: A multilingual, multifaceted dataset for summarization evaluation.
\newblock In \emph{Proceedings of the 2023 Conference on Empirical Methods in Natural Language Processing}.

\bibitem[{Clark et~al.(2020)Clark, Choi, Collins, Garrette, Kwiatkowski, Nikolaev, and Palomaki}]{clark2020tydi}
Jonathan~H Clark, Eunsol Choi, Michael Collins, Dan Garrette, Tom Kwiatkowski, Vitaly Nikolaev, and Jennimaria Palomaki. 2020.
\newblock Tydi qa: A benchmark for information-seeking question answering in ty pologically di verse languages.
\newblock \emph{Transactions of the Association for Computational Linguistics}, 8:454--470.

\bibitem[{Cobbe et~al.(2021)Cobbe, Kosaraju, Bavarian, Chen, Jun, Kaiser, Plappert, Tworek, Hilton, Nakano, Hesse, and Schulman}]{cobbe2021training}
Karl Cobbe, Vineet Kosaraju, Mohammad Bavarian, Mark Chen, Heewoo Jun, Lukasz Kaiser, Matthias Plappert, Jerry Tworek, Jacob Hilton, Reiichiro Nakano, Christopher Hesse, and John Schulman. 2021.
\newblock Training verifiers to solve math word problems.
\newblock \emph{arXiv preprint arXiv:2110.14168}.

\bibitem[{Cole et~al.(2023)Cole, Zhang, Gillick, Eisenschlos, Dhingra, and Eisenstein}]{cole-etal-2023-selectively}
Jeremy Cole, Michael Zhang, Daniel Gillick, Julian Eisenschlos, Bhuwan Dhingra, and Jacob Eisenstein. 2023.
\newblock Selectively answering ambiguous questions.
\newblock In \emph{Proceedings of the 2023 Conference on Empirical Methods in Natural Language Processing}.

\bibitem[{Conneau and Lample(2019)}]{conneau2019cross}
Alexis Conneau and Guillaume Lample. 2019.
\newblock Cross-lingual language model pretraining.
\newblock \emph{Advances in neural information processing systems}, 32.

\bibitem[{Conneau et~al.(2018)Conneau, Rinott, Lample, Williams, Bowman, Schwenk, and Stoyanov}]{conneau2018xnli}
Alexis Conneau, Ruty Rinott, Guillaume Lample, Adina Williams, Samuel Bowman, Holger Schwenk, and Veselin Stoyanov. 2018.
\newblock Xnli: Evaluating cross-lingual sentence representations.
\newblock In \emph{Proceedings of the 2018 Conference on Empirical Methods in Natural Language Processing}, pages 2475--2485.

\bibitem[{Dale et~al.(2023{\natexlab{a}})Dale, Voita, Barrault, and Costa-juss{\`a}}]{dale2023detecting}
David Dale, Elena Voita, Lo{\"\i}c Barrault, and Marta~R Costa-juss{\`a}. 2023{\natexlab{a}}.
\newblock Detecting and mitigating hallucinations in machine translation: Model internal workings alone do well, sentence similarity even better.
\newblock In \emph{Proceedings of the 61st Annual Meeting of the Association for Computational Linguistics (Volume 1: Long Papers)}, pages 36--50.

\bibitem[{Dale et~al.(2023{\natexlab{b}})Dale, Voita, Lam, Hansanti, Ropers, Kalbassi, Gao, Barrault, and Costa-juss{\`a}}]{dale-etal-2023-halomi}
David Dale, Elena Voita, Janice Lam, Prangthip Hansanti, Christophe Ropers, Elahe Kalbassi, Cynthia Gao, Loic Barrault, and Marta Costa-juss{\`a}. 2023{\natexlab{b}}.
\newblock {H}al{O}mi: A manually annotated benchmark for multilingual hallucination and omission detection in machine translation.
\newblock In \emph{Proceedings of the 2023 Conference on Empirical Methods in Natural Language Processing}.

\bibitem[{Desai and Durrett(2020)}]{desai-durrett-2020-calibration}
Shrey Desai and Greg Durrett. 2020.
\newblock Calibration of pre-trained transformers.
\newblock In \emph{Proceedings of the 2020 Conference on Empirical Methods in Natural Language Processing (EMNLP)}, pages 295--302, Online. Association for Computational Linguistics.

\bibitem[{Du et~al.(2023)Du, Li, Torralba, Tenenbaum, and Mordatch}]{du2023improving}
Yilun Du, Shuang Li, Antonio Torralba, Joshua~B Tenenbaum, and Igor Mordatch. 2023.
\newblock Improving factuality and reasoning in language models through multiagent debate.
\newblock \emph{arXiv preprint arXiv:2305.14325}.

\bibitem[{Ebrahimi et~al.(2022)Ebrahimi, Mager, Oncevay, Chaudhary, Chiruzzo, Fan, Ortega, Ramos, Rios, Meza~Ruiz, Gim{\'e}nez-Lugo, Mager, Neubig, Palmer, Coto-Solano, Vu, and Kann}]{ebrahimi-etal-2022-americasnli}
Abteen Ebrahimi, Manuel Mager, Arturo Oncevay, Vishrav Chaudhary, Luis Chiruzzo, Angela Fan, John Ortega, Ricardo Ramos, Annette Rios, Ivan~Vladimir Meza~Ruiz, Gustavo Gim{\'e}nez-Lugo, Elisabeth Mager, Graham Neubig, Alexis Palmer, Rolando Coto-Solano, Thang Vu, and Katharina Kann. 2022.
\newblock {A}mericas{NLI}: Evaluating zero-shot natural language understanding of pretrained multilingual models in truly low-resource languages.
\newblock In \emph{Proceedings of the 60th Annual Meeting of the Association for Computational Linguistics (Volume 1: Long Papers)}.

\bibitem[{Edunov et~al.(2018)Edunov, Ott, Auli, and Grangier}]{edunov2018understanding}
Sergey Edunov, Myle Ott, Michael Auli, and David Grangier. 2018.
\newblock Understanding back-translation at scale.
\newblock In \emph{Proceedings of the 2018 Conference on Empirical Methods in Natural Language Processing}, pages 489--500.

\bibitem[{Feng et~al.(2023)Feng, Shi, Bai, Balachandran, He, and Tsvetkov}]{Feng2023KnowledgeCF}
Shangbin Feng, Weijia Shi, Yuyang Bai, Vidhisha Balachandran, Tianxing He, and Yulia Tsvetkov. 2023.
\newblock Knowledge card: Filling llms' knowledge gaps with plug-in specialized language models.
\newblock In \emph{The Twelfth International Conference on Learning Representations}.

\bibitem[{Feng et~al.(2024)Feng, Shi, Wang, Ding, Balachandran, and Tsvetkov}]{feng2024don}
Shangbin Feng, Weijia Shi, Yike Wang, Wenxuan Ding, Vidhisha Balachandran, and Yulia Tsvetkov. 2024.
\newblock Don't hallucinate, abstain: Identifying llm knowledge gaps via multi-llm collaboration.
\newblock \emph{arXiv preprint arXiv:2402.00367}.

\bibitem[{Fujinuma et~al.(2022)Fujinuma, Boyd-Graber, and Kann}]{fujinuma-etal-2022-match}
Yoshinari Fujinuma, Jordan Boyd-Graber, and Katharina Kann. 2022.
\newblock Match the script, adapt if multilingual: Analyzing the effect of multilingual pretraining on cross-lingual transferability.
\newblock In \emph{Proceedings of the 60th Annual Meeting of the Association for Computational Linguistics (Volume 1: Long Papers)}.

\bibitem[{Gao et~al.(2024)Gao, Hu, Hu, Chen, Li, and Huang}]{gao-etal-2024-multilingual}
Changjiang Gao, Hongda Hu, Peng Hu, Jiajun Chen, Jixing Li, and Shujian Huang. 2024.
\newblock Multilingual pretraining and instruction tuning improve cross-lingual knowledge alignment, but only shallowly.
\newblock In \emph{Proceedings of the 2024 Conference of the North American Chapter of the Association for Computational Linguistics: Human Language Technologies (Volume 1: Long Papers)}.

\bibitem[{Gehrmann et~al.(2022)Gehrmann, Bhattacharjee, Mahendiran, Wang, Papangelis, Madaan, Mcmillan-major, Shvets, Upadhyay, Bohnet, Yao, Wilie, Bhagavatula, You, Thomson, Garbacea, Wang, Deutsch, Xiong, Jin, Gkatzia, Radev, Clark, Durmus, Ladhak, Ginter, Winata, Strobelt, Hayashi, Novikova, Kanerva, Chim, Zhou, Clive, Maynez, Sedoc, Juraska, Dhole, Chandu, Beltrachini, Ribeiro, Tunstall, Zhang, Pushkarna, Creutz, White, Kale, Eddine, Daheim, Subramani, Dusek, Liang, Ammanamanchi, Zhu, Puduppully, Kriz, Shahriyar, Cardenas, Mahamood, Osei, Cahyawijaya, {\v{S}}tajner, Montella, Jolly, Mille, Hasan, Shen, Adewumi, Raunak, Raheja, Nikolaev, Tsai, Jernite, Xu, Sang, Liu, and Hou}]{gehrmann-etal-2022-gemv2}
Sebastian Gehrmann, Abhik Bhattacharjee, Abinaya Mahendiran, Alex Wang, Alexandros Papangelis, Aman Madaan, Angelina Mcmillan-major, Anna Shvets, Ashish Upadhyay, Bernd Bohnet, Bingsheng Yao, Bryan Wilie, Chandra Bhagavatula, Chaobin You, Craig Thomson, Cristina Garbacea, Dakuo Wang, Daniel Deutsch, Deyi Xiong, Di~Jin, Dimitra Gkatzia, Dragomir Radev, Elizabeth Clark, Esin Durmus, Faisal Ladhak, Filip Ginter, Genta~Indra Winata, Hendrik Strobelt, Hiroaki Hayashi, Jekaterina Novikova, Jenna Kanerva, Jenny Chim, Jiawei Zhou, Jordan Clive, Joshua Maynez, Jo{\~a}o Sedoc, Juraj Juraska, Kaustubh Dhole, Khyathi~Raghavi Chandu, Laura~Perez Beltrachini, Leonardo F .~R. Ribeiro, Lewis Tunstall, Li~Zhang, Mahim Pushkarna, Mathias Creutz, Michael White, Mihir~Sanjay Kale, Moussa~Kamal Eddine, Nico Daheim, Nishant Subramani, Ondrej Dusek, Paul~Pu Liang, Pawan~Sasanka Ammanamanchi, Qi~Zhu, Ratish Puduppully, Reno Kriz, Rifat Shahriyar, Ronald Cardenas, Saad Mahamood, Salomey Osei, Samuel Cahyawijaya, Sanja {\v{S}}tajner,
  Sebastien Montella, Shailza Jolly, Simon Mille, Tahmid Hasan, Tianhao Shen, Tosin Adewumi, Vikas Raunak, Vipul Raheja, Vitaly Nikolaev, Vivian Tsai, Yacine Jernite, Ying Xu, Yisi Sang, Yixin Liu, and Yufang Hou. 2022.
\newblock {GEM}v2: Multilingual {NLG} benchmarking in a single line of code.
\newblock In \emph{Proceedings of the 2022 Conference on Empirical Methods in Natural Language Processing: System Demonstrations}.

\bibitem[{Gu and Hopkins(2023)}]{gu-hopkins-2023-evaluation}
Zhengyao Gu and Mark Hopkins. 2023.
\newblock On the evaluation of neural selective prediction methods for natural language processing.
\newblock In \emph{Proceedings of the 61st Annual Meeting of the Association for Computational Linguistics (Volume 1: Long Papers)}.

\bibitem[{He et~al.(2023)He, Cui, Chen, Hu, and Zhu}]{he2023investigating}
Guande He, Peng Cui, Jianfei Chen, Wenbo Hu, and Jun Zhu. 2023.
\newblock Investigating uncertainty calibration of aligned language models under the multiple-choice setting.
\newblock \emph{arXiv preprint arXiv:2310.11732}.

\bibitem[{Hu et~al.(2020)Hu, Ruder, Siddhant, Neubig, Firat, and Johnson}]{hu2020xtreme}
Junjie Hu, Sebastian Ruder, Aditya Siddhant, Graham Neubig, Orhan Firat, and Melvin Johnson. 2020.
\newblock Xtreme: A massively multilingual multi-task benchmark for evaluating cross-lingual generalisation.
\newblock In \emph{International Conference on Machine Learning}, pages 4411--4421. PMLR.

\bibitem[{Huang et~al.(2023{\natexlab{a}})Huang, Chen, Mishra, Zheng, Yu, Song, and Zhou}]{huang2023large}
Jie Huang, Xinyun Chen, Swaroop Mishra, Huaixiu~Steven Zheng, Adams~Wei Yu, Xinying Song, and Denny Zhou. 2023{\natexlab{a}}.
\newblock Large language models cannot self-correct reasoning yet.
\newblock In \emph{The Twelfth International Conference on Learning Representations}.

\bibitem[{Huang et~al.(2023{\natexlab{b}})Huang, Ruan, Huang, Jin, Dong, Wu, Bensalem, Mu, Qi, Zhao, Cai, Zhang, Wu, Xu, Wu, Freitas, and Mustafa}]{huang2023survey}
Xiaowei Huang, Wenjie Ruan, Wei Huang, Gao Jin, Yizhen Dong, Changshun Wu, Saddek Bensalem, Ronghui Mu, Yi~Qi, Xingyu Zhao, Kaiwen Cai, Yanghao Zhang, Sihao Wu, Peipei Xu, Dengyu Wu, Andr{\'e} Freitas, and Mustafa~A. Mustafa. 2023{\natexlab{b}}.
\newblock A survey of safety and trustworthiness of large language models through the lens of verification and validation.
\newblock \emph{ArXiv}, abs/2305.11391.

\bibitem[{Huang et~al.(2023{\natexlab{c}})Huang, Yu, and Allan}]{huang2023improving}
Zhiqi Huang, Puxuan Yu, and James Allan. 2023{\natexlab{c}}.
\newblock Improving cross-lingual information retrieval on low-resource languages via optimal transport distillation.
\newblock In \emph{Proceedings of the Sixteenth ACM International Conference on Web Search and Data Mining}, pages 1048--1056.

\bibitem[{Jagannatha and Yu(2020)}]{jagannatha-yu-2020-calibrating}
Abhyuday Jagannatha and Hong Yu. 2020.
\newblock Calibrating structured output predictors for natural language processing.
\newblock In \emph{Proceedings of the 58th Annual Meeting of the Association for Computational Linguistics}.

\bibitem[{Ji et~al.(2023)Ji, Lee, Frieske, Yu, Su, Xu, Ishii, Bang, Madotto, and Fung}]{ji2023survey}
Ziwei Ji, Nayeon Lee, Rita Frieske, Tiezheng Yu, Dan Su, Yan Xu, Etsuko Ishii, Ye~Jin Bang, Andrea Madotto, and Pascale Fung. 2023.
\newblock Survey of hallucination in natural language generation.
\newblock \emph{ACM Computing Surveys}, 55(12):1--38.

\bibitem[{Jiang et~al.(2021)Jiang, Araki, Ding, and Neubig}]{jiang2021can}
Zhengbao Jiang, Jun Araki, Haibo Ding, and Graham Neubig. 2021.
\newblock How can we know when language models know? on the calibration of language models for question answering.
\newblock \emph{Transactions of the Association for Computational Linguistics}, 9:962--977.

\bibitem[{Kadavath et~al.(2022)Kadavath, Conerly, Askell, Henighan, Drain, Perez, Schiefer, Dodds, DasSarma, Tran-Johnson, Johnston, El-Showk, Jones, Elhage, Hume, Chen, Bai, Bowman, Fort, Ganguli, Hernandez, Jacobson, Kernion, Kravec, Lovitt, Ndousse, Olsson, Ringer, Amodei, Brown, Clark, Joseph, Mann, McCandlish, Olah, and Kaplan}]{kadavath2022language}
Saurav Kadavath, Tom Conerly, Amanda Askell, T.~J. Henighan, Dawn Drain, Ethan Perez, Nicholas Schiefer, Zachary Dodds, Nova DasSarma, Eli Tran-Johnson, Scott Johnston, Sheer El-Showk, Andy Jones, Nelson Elhage, Tristan Hume, Anna Chen, Yuntao Bai, Sam Bowman, Stanislav Fort, Deep Ganguli, Danny Hernandez, Josh Jacobson, John Kernion, Shauna Kravec, Liane Lovitt, Kamal Ndousse, Catherine Olsson, Sam Ringer, Dario Amodei, Tom~B. Brown, Jack Clark, Nicholas Joseph, Benjamin Mann, Sam McCandlish, Christopher Olah, and Jared Kaplan. 2022.
\newblock Language models (mostly) know what they know.
\newblock \emph{ArXiv}, abs/2207.05221.

\bibitem[{Kalai and Vempala(2023)}]{kalai2023calibrated}
Adam~Tauman Kalai and Santosh~S Vempala. 2023.
\newblock Calibrated language models must hallucinate.
\newblock \emph{arXiv preprint arXiv:2311.14648}.

\bibitem[{Kamath et~al.(2020)Kamath, Jia, and Liang}]{kamath-etal-2020-selective}
Amita Kamath, Robin Jia, and Percy Liang. 2020.
\newblock Selective question answering under domain shift.
\newblock In \emph{Proceedings of the 58th Annual Meeting of the Association for Computational Linguistics}, pages 5684--5696, Online. Association for Computational Linguistics.

\bibitem[{Kang et~al.(2024)Kang, Blevins, and Zettlemoyer}]{kang2024comparing}
Haoqiang Kang, Terra Blevins, and Luke Zettlemoyer. 2024.
\newblock Comparing hallucination detection metrics for multilingual generation.
\newblock \emph{arXiv preprint arXiv:2402.10496}.

\bibitem[{Karthikeyan et~al.(2019)Karthikeyan, Wang, Mayhew, and Roth}]{karthikeyan2019cross}
K~Karthikeyan, Zihan Wang, Stephen Mayhew, and Dan Roth. 2019.
\newblock Cross-lingual ability of multilingual bert: An empirical study.
\newblock In \emph{International Conference on Learning Representations}.

\bibitem[{Kim et~al.(2024)Kim, Baldi, and McAleer}]{kim2023language}
Geunwoo Kim, Pierre Baldi, and Stephen McAleer. 2024.
\newblock Language models can solve computer tasks.
\newblock \emph{Advances in Neural Information Processing Systems}, 36.

\bibitem[{Kong et~al.(2020)Kong, Jiang, Zhuang, Lyu, Zhao, and Zhang}]{kong-etal-2020-calibrated}
Lingkai Kong, Haoming Jiang, Yuchen Zhuang, Jie Lyu, Tuo Zhao, and Chao Zhang. 2020.
\newblock Calibrated language model fine-tuning for in- and out-of-distribution data.
\newblock In \emph{Proceedings of the 2020 Conference on Empirical Methods in Natural Language Processing (EMNLP)}, pages 1326--1340, Online. Association for Computational Linguistics.

\bibitem[{Kuhn et~al.(2022)Kuhn, Gal, and Farquhar}]{kuhn2023semantic}
Lorenz Kuhn, Yarin Gal, and Sebastian Farquhar. 2022.
\newblock Semantic uncertainty: Linguistic invariances for uncertainty estimation in natural language generation.
\newblock In \emph{The Eleventh International Conference on Learning Representations}.

\bibitem[{Kumar et~al.(2023)Kumar, Balachandran, Njoo, Anastasopoulos, and Tsvetkov}]{kumar2023language}
Sachin Kumar, Vidhisha Balachandran, Lucille Njoo, Antonios Anastasopoulos, and Yulia Tsvetkov. 2023.
\newblock Language generation models can cause harm: So what can we do about it? an actionable survey.
\newblock In \emph{Proceedings of the 17th Conference of the European Chapter of the Association for Computational Linguistics}, pages 3299--3321.

\bibitem[{Lai et~al.(2023)Lai, Nguyen, Ngo, Nguyen, Dernoncourt, Rossi, and Nguyen}]{lai2023okapi}
Viet Lai, Chien Nguyen, Nghia Ngo, Thuat Nguyen, Franck Dernoncourt, Ryan Rossi, and Thien Nguyen. 2023.
\newblock Okapi: Instruction-tuned large language models in multiple languages with reinforcement learning from human feedback.
\newblock In \emph{Proceedings of the 2023 Conference on Empirical Methods in Natural Language Processing: System Demonstrations}, pages 318--327.

\bibitem[{Lazaridou et~al.(2021)Lazaridou, Kuncoro, Gribovskaya, Agrawal, Liska, Terzi, Gimenez, de~Masson~d'Autume, Kocisky, Ruder et~al.}]{lazaridou2021mind}
Angeliki Lazaridou, Adhi Kuncoro, Elena Gribovskaya, Devang Agrawal, Adam Liska, Tayfun Terzi, Mai Gimenez, Cyprien de~Masson~d'Autume, Tomas Kocisky, Sebastian Ruder, et~al. 2021.
\newblock Mind the gap: Assessing temporal generalization in neural language models.
\newblock \emph{Advances in Neural Information Processing Systems}, 34:29348--29363.

\bibitem[{Lee et~al.(2018)Lee, Firat, Agarwal, Fannjiang, and Sussillo}]{lee2018hallucinations}
Katherine Lee, Orhan Firat, Ashish Agarwal, Clara Fannjiang, and David Sussillo. 2018.
\newblock Hallucinations in neural machine translation.

\bibitem[{Lewis et~al.(2020)Lewis, Perez, Piktus, Petroni, Karpukhin, Goyal, K{\"u}ttler, Lewis, Yih, Rockt{\"a}schel et~al.}]{lewis2020retrieval}
Patrick Lewis, Ethan Perez, Aleksandra Piktus, Fabio Petroni, Vladimir Karpukhin, Naman Goyal, Heinrich K{\"u}ttler, Mike Lewis, Wen-tau Yih, Tim Rockt{\"a}schel, et~al. 2020.
\newblock Retrieval-augmented generation for knowledge-intensive nlp tasks.
\newblock \emph{NeurIPS}.

\bibitem[{Li et~al.(2024{\natexlab{a}})Li, Teney, Yang, Wen, Xie, and Wang}]{li2024culturepark}
Cheng Li, Damien Teney, Linyi Yang, Qingsong Wen, Xing Xie, and Jindong Wang. 2024{\natexlab{a}}.
\newblock Culturepark: Boosting cross-cultural understanding in large language models.
\newblock \emph{arXiv preprint arXiv:2405.15145}.

\bibitem[{Li et~al.(2024{\natexlab{b}})Li, Wang, Zhang, and Zong}]{li-etal-2024-improving-context}
Chong Li, Shaonan Wang, Jiajun Zhang, and Chengqing Zong. 2024{\natexlab{b}}.
\newblock Improving in-context learning of multilingual generative language models with cross-lingual alignment.
\newblock In \emph{Proceedings of the 2024 Conference of the North American Chapter of the Association for Computational Linguistics: Human Language Technologies (Volume 1: Long Papers)}.

\bibitem[{Li et~al.(2022)Li, Ding, Zhang, Cheng, Hu, and Luo}]{li-etal-2022-multi-level}
Mingqi Li, Fei Ding, Dan Zhang, Long Cheng, Hongxin Hu, and Feng Luo. 2022.
\newblock Multi-level distillation of semantic knowledge for pre-training multilingual language model.
\newblock In \emph{Proceedings of the 2022 Conference on Empirical Methods in Natural Language Processing}.

\bibitem[{Liang et~al.(2023)Liang, Bommasani, Lee, Tsipras, Soylu, Yasunaga, Zhang, Narayanan, Wu, Kumar et~al.}]{liang2022holistic}
Percy Liang, Rishi Bommasani, Tony Lee, Dimitris Tsipras, Dilara Soylu, Michihiro Yasunaga, Yian Zhang, Deepak Narayanan, Yuhuai Wu, Ananya Kumar, et~al. 2023.
\newblock Holistic evaluation of language models.
\newblock \emph{Transactions on Machine Learning Research}.

\bibitem[{Lin et~al.(2023)Lin, Ahmad, and Lin}]{lin-etal-2023-maggretriever}
Sheng-Chieh Lin, Amin Ahmad, and Jimmy Lin. 2023.
\newblock m{A}ggretriever: A simple yet effective approach to zero-shot multilingual dense retrieval.
\newblock In \emph{Proceedings of the 2023 Conference on Empirical Methods in Natural Language Processing}.

\bibitem[{Lin et~al.(2022{\natexlab{a}})Lin, Hilton, and Evans}]{lin2022teaching}
Stephanie Lin, Jacob Hilton, and Owain Evans. 2022{\natexlab{a}}.
\newblock Teaching models to express their uncertainty in words.
\newblock \emph{arXiv preprint arXiv:2205.14334}.

\bibitem[{Lin et~al.(2022{\natexlab{b}})Lin, Mihaylov, Artetxe, Wang, Chen, Simig, Ott, Goyal, Bhosale, Du, Pasunuru, Shleifer, Koura, Chaudhary, O{'}Horo, Wang, Zettlemoyer, Kozareva, Diab, Stoyanov, and Li}]{lin-etal-2022-shot}
Xi~Victoria Lin, Todor Mihaylov, Mikel Artetxe, Tianlu Wang, Shuohui Chen, Daniel Simig, Myle Ott, Naman Goyal, Shruti Bhosale, Jingfei Du, Ramakanth Pasunuru, Sam Shleifer, Punit~Singh Koura, Vishrav Chaudhary, Brian O{'}Horo, Jeff Wang, Luke Zettlemoyer, Zornitsa Kozareva, Mona Diab, Veselin Stoyanov, and Xian Li. 2022{\natexlab{b}}.
\newblock Few-shot learning with multilingual generative language models.
\newblock In \emph{Proceedings of the 2022 Conference on Empirical Methods in Natural Language Processing}.

\bibitem[{Lin et~al.(2019{\natexlab{a}})Lin, Chen, Lee, Li, Zhang, Xia, Rijhwani, He, Zhang, Ma, Anastasopoulos, Littell, and Neubig}]{lin-etal-2019-choosing}
Yu-Hsiang Lin, Chian-Yu Chen, Jean Lee, Zirui Li, Yuyan Zhang, Mengzhou Xia, Shruti Rijhwani, Junxian He, Zhisong Zhang, Xuezhe Ma, Antonios Anastasopoulos, Patrick Littell, and Graham Neubig. 2019{\natexlab{a}}.
\newblock Choosing transfer languages for cross-lingual learning.
\newblock In \emph{Proceedings of the 57th Annual Meeting of the Association for Computational Linguistics}.

\bibitem[{Lin et~al.(2019{\natexlab{b}})Lin, Chen, Lee, Li, Zhang, Xia, Rijhwani, He, Zhang, Ma et~al.}]{lin2019choosing}
Yu-Hsiang Lin, Chian-Yu Chen, Jean Lee, Zirui Li, Yuyan Zhang, Mengzhou Xia, Shruti Rijhwani, Junxian He, Zhisong Zhang, Xuezhe Ma, et~al. 2019{\natexlab{b}}.
\newblock Choosing transfer languages for cross-lingual learning.
\newblock In \emph{Proceedings of the 57th Annual Meeting of the Association for Computational Linguistics}, pages 3125--3135.

\bibitem[{Littell et~al.(2017)Littell, Mortensen, Lin, Kairis, Turner, and Levin}]{littell2017uriel}
Patrick Littell, David~R Mortensen, Ke~Lin, Katherine Kairis, Carlisle Turner, and Lori Levin. 2017.
\newblock Uriel and lang2vec: Representing languages as typological, geographical, and phylogenetic vectors.
\newblock In \emph{Proceedings of the 15th Conference of the European Chapter of the Association for Computational Linguistics: Volume 2, Short Papers}, pages 8--14.

\bibitem[{Liu et~al.(2022)Liu, Li, He, Bing, Joty, and Si}]{liu-etal-2022-enhancing-multilingual}
Linlin Liu, Xin Li, Ruidan He, Lidong Bing, Shafiq Joty, and Luo Si. 2022.
\newblock Enhancing multilingual language model with massive multilingual knowledge triples.
\newblock In \emph{Proceedings of the 2022 Conference on Empirical Methods in Natural Language Processing}.

\bibitem[{Liu et~al.(2023{\natexlab{a}})Liu, Khalifa, and Wang}]{liu2023litcab}
Xin Liu, Muhammad Khalifa, and Lu~Wang. 2023{\natexlab{a}}.
\newblock Litcab: Lightweight calibration of language models on outputs of varied lengths.
\newblock \emph{arXiv preprint arXiv:2310.19208}.

\bibitem[{Liu et~al.(2023{\natexlab{b}})Liu, Yao, Ton, Zhang, Guo, Cheng, Klochkov, Taufiq, and Li}]{liu2023trustworthy}
Yang Liu, Yuanshun Yao, Jean-Francois Ton, Xiaoying Zhang, Ruocheng Guo, Hao Cheng, Yegor Klochkov, Muhammad~Faaiz Taufiq, and Hang Li. 2023{\natexlab{b}}.
\newblock Trustworthy llms: a survey and guideline for evaluating large language models' alignment.
\newblock In \emph{Socially Responsible Language Modelling Research}.

\bibitem[{Longpre et~al.(2021)Longpre, Lu, and Daiber}]{longpre2021mkqa}
Shayne Longpre, Yi~Lu, and Joachim Daiber. 2021.
\newblock Mkqa: A linguistically diverse benchmark for multilingual open domain question answering.
\newblock \emph{Transactions of the Association for Computational Linguistics}, 9:1389--1406.

\bibitem[{Madaan et~al.(2024)Madaan, Tandon, Gupta, Hallinan, Gao, Wiegreffe, Alon, Dziri, Prabhumoye, Yang et~al.}]{madaan2024self}
Aman Madaan, Niket Tandon, Prakhar Gupta, Skyler Hallinan, Luyu Gao, Sarah Wiegreffe, Uri Alon, Nouha Dziri, Shrimai Prabhumoye, Yiming Yang, et~al. 2024.
\newblock Self-refine: Iterative refinement with self-feedback.
\newblock \emph{Advances in Neural Information Processing Systems}, 36.

\bibitem[{Mielke et~al.(2022)Mielke, Szlam, Dinan, and Boureau}]{mielke2022reducing}
Sabrina~J Mielke, Arthur Szlam, Emily Dinan, and Y-Lan Boureau. 2022.
\newblock Reducing conversational agents’ overconfidence through linguistic calibration.
\newblock \emph{Transactions of the Association for Computational Linguistics}, 10:857--872.

\bibitem[{Mishra et~al.(2024)Mishra, Asai, Balachandran, Wang, Neubig, Tsvetkov, and Hajishirzi}]{mishra2024fine}
Abhika Mishra, Akari Asai, Vidhisha Balachandran, Yizhong Wang, Graham Neubig, Yulia Tsvetkov, and Hannaneh Hajishirzi. 2024.
\newblock Fine-grained hallucination detection and editing for language models.
\newblock \emph{arXiv preprint arXiv:2401.06855}.

\bibitem[{Muennighoff et~al.(2023)Muennighoff, Wang, Sutawika, Roberts, Biderman, Le~Scao, Bari, Shen, Yong, Schoelkopf et~al.}]{muennighoff2023crosslingual}
Niklas Muennighoff, Thomas Wang, Lintang Sutawika, Adam Roberts, Stella Biderman, Teven Le~Scao, M~Saiful Bari, Sheng Shen, Zheng~Xin Yong, Hailey Schoelkopf, et~al. 2023.
\newblock Crosslingual generalization through multitask finetuning.
\newblock In \emph{Proceedings of the 61st Annual Meeting of the Association for Computational Linguistics (Volume 1: Long Papers)}, pages 15991--16111.

\bibitem[{Naous et~al.(2023)Naous, Ryan, Ritter, and Xu}]{naous2023having}
Tarek Naous, Michael~J Ryan, Alan Ritter, and Wei Xu. 2023.
\newblock Having beer after prayer? measuring cultural bias in large language models.
\newblock \emph{arXiv preprint arXiv:2305.14456}.

\bibitem[{Ogundepo et~al.(2023)Ogundepo, Gwadabe, Rivera, Clark, Ruder, Adelani, Dossou, Diop, Sikasote, Hacheme et~al.}]{ogundepo2023cross}
Odunayo Ogundepo, Tajuddeen Gwadabe, Clara Rivera, Jonathan~H Clark, Sebastian Ruder, David Adelani, Bonaventure Dossou, Abdou Diop, Claytone Sikasote, Gilles Hacheme, et~al. 2023.
\newblock Cross-lingual open-retrieval question answering for african languages.
\newblock In \emph{Findings of the Association for Computational Linguistics: EMNLP 2023}, pages 14957--14972.

\bibitem[{Ouyang et~al.(2022)Ouyang, Wu, Jiang, Almeida, Wainwright, Mishkin, Zhang, Agarwal, Slama, Ray et~al.}]{ouyang2022training}
Long Ouyang, Jeffrey Wu, Xu~Jiang, Diogo Almeida, Carroll Wainwright, Pamela Mishkin, Chong Zhang, Sandhini Agarwal, Katarina Slama, Alex Ray, et~al. 2022.
\newblock Training language models to follow instructions with human feedback.
\newblock \emph{Advances in Neural Information Processing Systems}, 35:27730--27744.

\bibitem[{Petroni et~al.(2019)Petroni, Rockt{\"a}schel, Riedel, Lewis, Bakhtin, Wu, and Miller}]{petroni2019language}
Fabio Petroni, Tim Rockt{\"a}schel, Sebastian Riedel, Patrick Lewis, Anton Bakhtin, Yuxiang Wu, and Alexander Miller. 2019.
\newblock Language models as knowledge bases?
\newblock In \emph{Proceedings of the 2019 Conference on Empirical Methods in Natural Language Processing and the 9th International Joint Conference on Natural Language Processing (EMNLP-IJCNLP)}, pages 2463--2473.

\bibitem[{Philippy et~al.(2023)Philippy, Guo, and Haddadan}]{philippy-etal-2023-towards}
Fred Philippy, Siwen Guo, and Shohreh Haddadan. 2023.
\newblock Towards a common understanding of contributing factors for cross-lingual transfer in multilingual language models: A review.
\newblock In \emph{Proceedings of the 61st Annual Meeting of the Association for Computational Linguistics (Volume 1: Long Papers)}.

\bibitem[{Pires et~al.(2019)Pires, Schlinger, and Garrette}]{pires-etal-2019-multilingual}
Telmo Pires, Eva Schlinger, and Dan Garrette. 2019.
\newblock How multilingual is multilingual {BERT}?
\newblock In \emph{Proceedings of the 57th Annual Meeting of the Association for Computational Linguistics}.

\bibitem[{Qi et~al.(2023)Qi, Fern{\'a}ndez, and Bisazza}]{qi-etal-2023-cross}
Jirui Qi, Raquel Fern{\'a}ndez, and Arianna Bisazza. 2023.
\newblock Cross-lingual consistency of factual knowledge in multilingual language models.
\newblock In \emph{Proceedings of the 2023 Conference on Empirical Methods in Natural Language Processing}.

\bibitem[{Qiu et~al.(2023)Qiu, Ziser, Korhonen, Ponti, and Cohen}]{qiu2023detecting}
Yifu Qiu, Yftah Ziser, Anna Korhonen, Edoardo Ponti, and Shay~B Cohen. 2023.
\newblock Detecting and mitigating hallucinations in multilingual summarisation.
\newblock In \emph{Proceedings of the 2023 Conference on Empirical Methods in Natural Language Processing}, pages 8914--8932.

\bibitem[{Radford et~al.(2019)Radford, Wu, Child, Luan, Amodei, Sutskever et~al.}]{radford2019language}
Alec Radford, Jeffrey Wu, Rewon Child, David Luan, Dario Amodei, Ilya Sutskever, et~al. 2019.
\newblock Language models are unsupervised multitask learners.
\newblock \emph{OpenAI blog}, 1(8):9.

\bibitem[{Rao et~al.(2024)Rao, Yerukola, Shah, Reinecke, and Sap}]{rao2024normad}
Abhinav Rao, Akhila Yerukola, Vishwa Shah, Katharina Reinecke, and Maarten Sap. 2024.
\newblock Normad: A benchmark for measuring the cultural adaptability of large language models.
\newblock \emph{arXiv preprint arXiv:2404.12464}.

\bibitem[{Raunak et~al.(2021)Raunak, Menezes, and Junczys-Dowmunt}]{raunak2021curious}
Vikas Raunak, Arul Menezes, and Marcin Junczys-Dowmunt. 2021.
\newblock The curious case of hallucinations in neural machine translation.
\newblock In \emph{Proceedings of the 2021 Conference of the North American Chapter of the Association for Computational Linguistics: Human Language Technologies}, pages 1172--1183.

\bibitem[{Reusens et~al.(2023)Reusens, Borchert, Mieskes, De~Weerdt, and Baesens}]{reusens-etal-2023-investigating}
Manon Reusens, Philipp Borchert, Margot Mieskes, Jochen De~Weerdt, and Bart Baesens. 2023.
\newblock Investigating bias in multilingual language models: Cross-lingual transfer of debiasing techniques.
\newblock In \emph{Proceedings of the 2023 Conference on Empirical Methods in Natural Language Processing}.

\bibitem[{Schmidt et~al.(2022)Schmidt, Vuli{\'c}, and Glava{\v{s}}}]{schmidt-etal-2022-dont}
Fabian~David Schmidt, Ivan Vuli{\'c}, and Goran Glava{\v{s}}. 2022.
\newblock Don{'}t stop fine-tuning: On training regimes for few-shot cross-lingual transfer with multilingual language models.
\newblock In \emph{Proceedings of the 2022 Conference on Empirical Methods in Natural Language Processing}.

\bibitem[{Schott et~al.(2023)Schott, Furman, and Bhat}]{schott-etal-2023-polyglot}
Tim Schott, Daniel Furman, and Shreshta Bhat. 2023.
\newblock Polyglot or not? measuring multilingual encyclopedic knowledge in foundation models.
\newblock In \emph{Proceedings of the 2023 Conference on Empirical Methods in Natural Language Processing}.

\bibitem[{Shen et~al.(2022)Shen, Liu, Zhou, and Xiong}]{shen-etal-2022-recovering}
Tianhao Shen, Mingtong Liu, Ming Zhou, and Deyi Xiong. 2022.
\newblock Recovering gold from black sand: Multilingual dense passage retrieval with hard and false negative samples.
\newblock In \emph{Proceedings of the 2022 Conference on Empirical Methods in Natural Language Processing}.

\bibitem[{Shi et~al.(2024)Shi, Min, Yasunaga, Seo, James, Lewis, Zettlemoyer, and Yih}]{shi-etal-2024-replug}
Weijia Shi, Sewon Min, Michihiro Yasunaga, Minjoon Seo, Richard James, Mike Lewis, Luke Zettlemoyer, and Wen-tau Yih. 2024.
\newblock {REPLUG}: Retrieval-augmented black-box language models.
\newblock In \emph{Proceedings of the 2024 Conference of the North American Chapter of the Association for Computational Linguistics: Human Language Technologies (Volume 1: Long Papers)}.

\bibitem[{Shinn et~al.(2023)Shinn, Cassano, Gopinath, Narasimhan, and Yao}]{shinn2023reflexion}
Noah Shinn, Federico Cassano, Ashwin Gopinath, Karthik~R Narasimhan, and Shunyu Yao. 2023.
\newblock Reflexion: Language agents with verbal reinforcement learning.
\newblock In \emph{Thirty-seventh Conference on Neural Information Processing Systems}.

\bibitem[{Si et~al.(2023)Si, Shi, Zhao, Zettlemoyer, and Boyd-Graber}]{si2023getting}
Chenglei Si, Weijia Shi, Chen Zhao, Luke Zettlemoyer, and Jordan Boyd-Graber. 2023.
\newblock Getting more out of mixture of language model reasoning experts.
\newblock In \emph{Findings of the Association for Computational Linguistics: EMNLP 2023}, pages 8234--8249.

\bibitem[{Slobodkin et~al.(2023)Slobodkin, Goldman, Caciularu, Dagan, and Ravfogel}]{slobodkin2023curious}
Aviv Slobodkin, Omer Goldman, Avi Caciularu, Ido Dagan, and Shauli Ravfogel. 2023.
\newblock The curious case of hallucinatory (un) answerability: Finding truths in the hidden states of over-confident large language models.
\newblock In \emph{Proceedings of the 2023 Conference on Empirical Methods in Natural Language Processing}, pages 3607--3625.

\bibitem[{Song et~al.(2023)Song, Khanuja, Liu, Faisal, Ostapenko, Winata, Aji, Cahyawijaya, Tsvetkov, Anastasopoulos et~al.}]{song2023globalbench}
Yueqi Song, Simran Khanuja, Pengfei Liu, Fahim Faisal, Alissa Ostapenko, Genta Winata, Alham Aji, Samuel Cahyawijaya, Yulia Tsvetkov, Antonios Anastasopoulos, et~al. 2023.
\newblock Globalbench: A benchmark for global progress in natural language processing.
\newblock In \emph{Proceedings of the 2023 Conference on Empirical Methods in Natural Language Processing}, pages 14157--14171.

\bibitem[{Stengel-Eskin and Van~Durme(2023)}]{stengel2023calibrated}
Elias Stengel-Eskin and Benjamin Van~Durme. 2023.
\newblock Calibrated interpretation: Confidence estimation in semantic parsing.
\newblock \emph{Transactions of the Association for Computational Linguistics}, 11:1213--1231.

\bibitem[{Sun et~al.(2021)Sun, Ahn, Park, Tsvetkov, and Mortensen}]{sun2021cross}
Jimin Sun, Hwijeen Ahn, Chan~Young Park, Yulia Tsvetkov, and David~R Mortensen. 2021.
\newblock Cross-cultural similarity features for cross-lingual transfer learning of pragmatically motivated tasks.
\newblock In \emph{The 16th Conference of the European Chapter of the Association for Computational Linguistics (EACL)}.

\bibitem[{Sun et~al.(2022)Sun, Yan, Abbeel, and Mordatch}]{sun2022quantifying}
Meiqi Sun, Wilson Yan, Pieter Abbeel, and Igor Mordatch. 2022.
\newblock Quantifying uncertainty in foundation models via ensembles.
\newblock In \emph{NeurIPS 2022 Workshop on Robustness in Sequence Modeling}.

\bibitem[{Sun et~al.(2023)Sun, Shen, Cao, Liu, Li, Shen, Gan, Gui, Wang, Yang et~al.}]{sun2023aligning}
Zhiqing Sun, Sheng Shen, Shengcao Cao, Haotian Liu, Chunyuan Li, Yikang Shen, Chuang Gan, Liang-Yan Gui, Yu-Xiong Wang, Yiming Yang, et~al. 2023.
\newblock Aligning large multimodal models with factually augmented rlhf.
\newblock \emph{arXiv preprint arXiv:2309.14525}.

\bibitem[{Tanwar et~al.(2023)Tanwar, Dutta, Borthakur, and Chakraborty}]{tanwar-etal-2023-multilingual}
Eshaan Tanwar, Subhabrata Dutta, Manish Borthakur, and Tanmoy Chakraborty. 2023.
\newblock Multilingual {LLM}s are better cross-lingual in-context learners with alignment.
\newblock In \emph{Proceedings of the 61st Annual Meeting of the Association for Computational Linguistics (Volume 1: Long Papers)}.

\bibitem[{Tao et~al.(2023)Tao, Zhu, Guo, Dong, and Xu}]{tao2023benchmark}
Linwei Tao, Younan Zhu, Haolan Guo, Minjing Dong, and Chang Xu. 2023.
\newblock A benchmark study on calibration.
\newblock In \emph{The Twelfth International Conference on Learning Representations}.

\bibitem[{Thakur et~al.(2024)Thakur, Ni, Hernandez~Abrego, Wieting, Lin, and Cer}]{thakur-etal-2024-leveraging}
Nandan Thakur, Jianmo Ni, Gustavo Hernandez~Abrego, John Wieting, Jimmy Lin, and Daniel Cer. 2024.
\newblock Leveraging {LLM}s for synthesizing training data across many languages in multilingual dense retrieval.
\newblock In \emph{Proceedings of the 2024 Conference of the North American Chapter of the Association for Computational Linguistics: Human Language Technologies (Volume 1: Long Papers)}.

\bibitem[{Tian et~al.(2023)Tian, Mitchell, Zhou, Sharma, Rafailov, Yao, Finn, and Manning}]{tian2023just}
Katherine Tian, Eric Mitchell, Allan Zhou, Archit Sharma, Rafael Rafailov, Huaxiu Yao, Chelsea Finn, and Christopher Manning. 2023.
\newblock Just ask for calibration: Strategies for eliciting calibrated confidence scores from language models fine-tuned with human feedback.
\newblock In \emph{Proceedings of the 2023 Conference on Empirical Methods in Natural Language Processing}, pages 5433--5442, Singapore. Association for Computational Linguistics.

\bibitem[{{\"U}st{\"u}n et~al.(2024){\"U}st{\"u}n, Aryabumi, Yong, Ko, D'souza, Onilude, Bhandari, Singh, Ooi, Kayid et~al.}]{ustun2024aya}
Ahmet {\"U}st{\"u}n, Viraat Aryabumi, Zheng-Xin Yong, Wei-Yin Ko, Daniel D'souza, Gbemileke Onilude, Neel Bhandari, Shivalika Singh, Hui-Lee Ooi, Amr Kayid, et~al. 2024.
\newblock Aya model: An instruction finetuned open-access multilingual language model.
\newblock \emph{arXiv preprint arXiv:2402.07827}.

\bibitem[{{\"U}st{\"u}n et~al.(2022){\"U}st{\"u}n, Bisazza, Bouma, van Noord, and Ruder}]{ustun-etal-2022-hyper}
Ahmet {\"U}st{\"u}n, Arianna Bisazza, Gosse Bouma, Gertjan van Noord, and Sebastian Ruder. 2022.
\newblock Hyper-{X}: A unified hypernetwork for multi-task multilingual transfer.
\newblock In \emph{Proceedings of the 2022 Conference on Empirical Methods in Natural Language Processing}.

\bibitem[{Varshney and Baral(2023)}]{varshney-baral-2023-post}
Neeraj Varshney and Chitta Baral. 2023.
\newblock Post-abstention: Towards reliably re-attempting the abstained instances in {QA}.
\newblock In \emph{Proceedings of the 61st Annual Meeting of the Association for Computational Linguistics (Volume 1: Long Papers)}.

\bibitem[{Wang et~al.(2024)Wang, Liu, Huang, Jiao, Ding, Aw, and Chen}]{wang-etal-2024-seaeval}
Bin Wang, Zhengyuan Liu, Xin Huang, Fangkai Jiao, Yang Ding, AiTi Aw, and Nancy Chen. 2024.
\newblock {S}ea{E}val for multilingual foundation models: From cross-lingual alignment to cultural reasoning.
\newblock In \emph{Proceedings of the 2024 Conference of the North American Chapter of the Association for Computational Linguistics: Human Language Technologies (Volume 1: Long Papers)}.

\bibitem[{Wang et~al.(2023{\natexlab{a}})Wang, Yue, and Sun}]{wang-etal-2023-chatgpt-defend}
Boshi Wang, Xiang Yue, and Huan Sun. 2023{\natexlab{a}}.
\newblock Can {C}hat{GPT} defend its belief in truth? evaluating {LLM} reasoning via debate.
\newblock In \emph{Findings of the Association for Computational Linguistics: EMNLP 2023}, pages 11865--11881, Singapore. Association for Computational Linguistics.

\bibitem[{Wang et~al.(2020{\natexlab{a}})Wang, Tu, Shi, and Liu}]{wang-etal-2020-inference}
Shuo Wang, Zhaopeng Tu, Shuming Shi, and Yang Liu. 2020{\natexlab{a}}.
\newblock On the inference calibration of neural machine translation.
\newblock In \emph{Proceedings of the 58th Annual Meeting of the Association for Computational Linguistics}, pages 3070--3079, Online. Association for Computational Linguistics.

\bibitem[{Wang et~al.(2022)Wang, Wei, Schuurmans, Le, Chi, Narang, Chowdhery, and Zhou}]{wang2022self}
Xuezhi Wang, Jason Wei, Dale Schuurmans, Quoc~V Le, Ed~H Chi, Sharan Narang, Aakanksha Chowdhery, and Denny Zhou. 2022.
\newblock Self-consistency improves chain of thought reasoning in language models.
\newblock In \emph{The Eleventh International Conference on Learning Representations}.

\bibitem[{Wang et~al.(2023{\natexlab{b}})Wang, Feng, Wang, Shi, Balachandran, He, and Tsvetkov}]{wang2023resolving}
Yike Wang, Shangbin Feng, Heng Wang, Weijia Shi, Vidhisha Balachandran, Tianxing He, and Yulia Tsvetkov. 2023{\natexlab{b}}.
\newblock Resolving knowledge conflicts in large language models.
\newblock \emph{arXiv preprint arXiv:2310.00935}.

\bibitem[{Wang et~al.(2020{\natexlab{b}})Wang, Lipton, and Tsvetkov}]{wang-etal-2020-negative}
Zirui Wang, Zachary~C. Lipton, and Yulia Tsvetkov. 2020{\natexlab{b}}.
\newblock On negative interference in multilingual models: Findings and a meta-learning treatment.
\newblock In \emph{Proceedings of the 2020 Conference on Empirical Methods in Natural Language Processing (EMNLP)}, pages 4438--4450. Association for Computational Linguistics.

\bibitem[{Whitehead et~al.(2022)Whitehead, Petryk, Shakib, Gonzalez, Darrell, Rohrbach, and Rohrbach}]{whitehead2022reliable}
Spencer Whitehead, Suzanne Petryk, Vedaad Shakib, Joseph Gonzalez, Trevor Darrell, Anna Rohrbach, and Marcus Rohrbach. 2022.
\newblock Reliable visual question answering: Abstain rather than answer incorrectly.
\newblock In \emph{European Conference on Computer Vision}, pages 148--166. Springer.

\bibitem[{Wieting et~al.(2023)Wieting, Clark, Cohen, Neubig, and Berg-Kirkpatrick}]{wieting-etal-2023-beyond}
John Wieting, Jonathan Clark, William Cohen, Graham Neubig, and Taylor Berg-Kirkpatrick. 2023.
\newblock Beyond contrastive learning: A variational generative model for multilingual retrieval.
\newblock In \emph{Proceedings of the 61st Annual Meeting of the Association for Computational Linguistics (Volume 1: Long Papers)}.

\bibitem[{Wu et~al.(2022)Wu, Wu, Zhang, Xiong, Chen, Zhuang, and Feng}]{wu-etal-2022-learning}
Linjuan Wu, Shaojuan Wu, Xiaowang Zhang, Deyi Xiong, Shizhan Chen, Zhiqiang Zhuang, and Zhiyong Feng. 2022.
\newblock Learning disentangled semantic representations for zero-shot cross-lingual transfer in multilingual machine reading comprehension.
\newblock In \emph{Proceedings of the 60th Annual Meeting of the Association for Computational Linguistics (Volume 1: Long Papers)}.

\bibitem[{Wu and Dredze(2019)}]{wu-dredze-2019-beto}
Shijie Wu and Mark Dredze. 2019.
\newblock Beto, bentz, becas: The surprising cross-lingual effectiveness of {BERT}.
\newblock In \emph{Proceedings of the 2019 Conference on Empirical Methods in Natural Language Processing and the 9th International Joint Conference on Natural Language Processing (EMNLP-IJCNLP)}.

\bibitem[{Xie et~al.(2023)Xie, Zhang, Chen, Lou, and Su}]{xie2023adaptive}
Jian Xie, Kai Zhang, Jiangjie Chen, Renze Lou, and Yu~Su. 2023.
\newblock Adaptive chameleon or stubborn sloth: Revealing the behavior of large language models in knowledge conflicts.
\newblock In \emph{The Twelfth International Conference on Learning Representations}.

\bibitem[{Xu et~al.(2024)Xu, Shi, and Choi}]{xu2024recomp}
Fangyuan Xu, Weijia Shi, and Eunsol Choi. 2024.
\newblock {RECOMP}: Improving retrieval-augmented {LM}s with context compression and selective augmentation.
\newblock In \emph{The Twelfth International Conference on Learning Representations}.

\bibitem[{Xu et~al.(2023{\natexlab{a}})Xu, Tan, Li, Chen, Van~Durme, Koehn, and Murray}]{xu-etal-2023-condensing}
Haoran Xu, Weiting Tan, Shuyue Li, Yunmo Chen, Benjamin Van~Durme, Philipp Koehn, and Kenton Murray. 2023{\natexlab{a}}.
\newblock Condensing multilingual knowledge with lightweight language-specific modules.
\newblock In \emph{Proceedings of the 2023 Conference on Empirical Methods in Natural Language Processing}.

\bibitem[{Xu et~al.(2023{\natexlab{b}})Xu, Li, and Xiong}]{xu-etal-2023-language-representation}
Shaoyang Xu, Junzhuo Li, and Deyi Xiong. 2023{\natexlab{b}}.
\newblock Language representation projection: Can we transfer factual knowledge across languages in multilingual language models?
\newblock In \emph{Proceedings of the 2023 Conference on Empirical Methods in Natural Language Processing}.

\bibitem[{Xu et~al.(2023{\natexlab{c}})Xu, Agrawal, Briakou, Martindale, and Carpuat}]{xu2023understanding}
Weijia Xu, Sweta Agrawal, Eleftheria Briakou, Marianna~J Martindale, and Marine Carpuat. 2023{\natexlab{c}}.
\newblock Understanding and detecting hallucinations in neural machine translation via model introspection.
\newblock \emph{Transactions of the Association for Computational Linguistics}, 11:546--564.

\bibitem[{Yang et~al.(2023)Yang, Chern, Qiu, Neubig, and Liu}]{yang2023alignment}
Yuqing Yang, Ethan Chern, Xipeng Qiu, Graham Neubig, and Pengfei Liu. 2023.
\newblock Alignment for honesty.
\newblock \emph{arXiv preprint arXiv:2312.07000}.

\bibitem[{Yasunaga et~al.(2023)Yasunaga, Aghajanyan, Shi, James, Leskovec, Liang, Lewis, Zettlemoyer, and Yih}]{yasunaga2023retrieval}
Michihiro Yasunaga, Armen Aghajanyan, Weijia Shi, Richard James, Jure Leskovec, Percy Liang, Mike Lewis, Luke Zettlemoyer, and Wen-tau Yih. 2023.
\newblock Retrieval-augmented multimodal language modeling.

\bibitem[{Yu et~al.(2023)Yu, Wang, Tu, Cao, Zhang-Li, Lv, Peng, Yao, Zhang, Li et~al.}]{yu2023kola}
Jifan Yu, Xiaozhi Wang, Shangqing Tu, Shulin Cao, Daniel Zhang-Li, Xin Lv, Hao Peng, Zijun Yao, Xiaohan Zhang, Hanming Li, et~al. 2023.
\newblock Kola: Carefully benchmarking world knowledge of large language models.
\newblock In \emph{The Twelfth International Conference on Learning Representations}.

\bibitem[{Zablotskaia et~al.(2023)Zablotskaia, Phan, Maynez, Narayan, Ren, and Liu}]{zablotskaia-etal-2023-uncertainty}
Polina Zablotskaia, Du~Phan, Joshua Maynez, Shashi Narayan, Jie Ren, and Jeremiah Liu. 2023.
\newblock On uncertainty calibration and selective generation in probabilistic neural summarization: A benchmark study.
\newblock In \emph{Findings of the Association for Computational Linguistics: EMNLP 2023}.

\bibitem[{Zhang et~al.(2023{\natexlab{a}})Zhang, Diao, Lin, Fung, Lian, Wang, Chen, Ji, and Zhang}]{zhang2023r}
Hanning Zhang, Shizhe Diao, Yong Lin, Yi~R Fung, Qing Lian, Xingyao Wang, Yangyi Chen, Heng Ji, and Tong Zhang. 2023{\natexlab{a}}.
\newblock R-tuning: Teaching large language models to refuse unknown questions.
\newblock \emph{arXiv preprint arXiv:2311.09677}.

\bibitem[{Zhang et~al.(2023{\natexlab{b}})Zhang, Li, Hauer, Shi, and Kondrak}]{zhang-etal-2023-dont}
Xiang Zhang, Senyu Li, Bradley Hauer, Ning Shi, and Grzegorz Kondrak. 2023{\natexlab{b}}.
\newblock Don{'}t trust {C}hat{GPT} when your question is not in {E}nglish: A study of multilingual abilities and types of {LLM}s.
\newblock In \emph{Proceedings of the 2023 Conference on Empirical Methods in Natural Language Processing}.

\bibitem[{Zhang et~al.(2023{\natexlab{c}})Zhang, Li, Cui, Cai, Liu, Fu, Huang, Zhao, Zhang, Chen et~al.}]{zhang2023siren}
Yue Zhang, Yafu Li, Leyang Cui, Deng Cai, Lemao Liu, Tingchen Fu, Xinting Huang, Enbo Zhao, Yu~Zhang, Yulong Chen, et~al. 2023{\natexlab{c}}.
\newblock Siren's song in the ai ocean: a survey on hallucination in large language models.
\newblock \emph{arXiv preprint arXiv:2309.01219}.

\bibitem[{Zheng et~al.(2024)Zheng, Chiang, Sheng, Zhuang, Wu, Zhuang, Lin, Li, Li, Xing et~al.}]{zheng2024judging}
Lianmin Zheng, Wei-Lin Chiang, Ying Sheng, Siyuan Zhuang, Zhanghao Wu, Yonghao Zhuang, Zi~Lin, Zhuohan Li, Dacheng Li, Eric Xing, et~al. 2024.
\newblock Judging llm-as-a-judge with mt-bench and chatbot arena.
\newblock \emph{Advances in Neural Information Processing Systems}, 36.

\bibitem[{Zhou et~al.(2023{\natexlab{a}})Zhou, Wan, Proleev, Mincu, Chen, Heller, and Roy}]{zhou2023batch}
Han Zhou, Xingchen Wan, Lev Proleev, Diana Mincu, Jilin Chen, Katherine Heller, and Subhrajit Roy. 2023{\natexlab{a}}.
\newblock Batch calibration: Rethinking calibration for in-context learning and prompt engineering.
\newblock \emph{arXiv preprint arXiv:2309.17249}.

\bibitem[{Zhou et~al.(2024)Zhou, Hwang, Ren, and Sap}]{Zhou2024RelyingOT}
Kaitlyn Zhou, Jena~D Hwang, Xiang Ren, and Maarten Sap. 2024.
\newblock Relying on the unreliable: The impact of language models' reluctance to express uncertainty.
\newblock \emph{arXiv preprint arXiv:2401.06730}.

\bibitem[{Zhou et~al.(2023{\natexlab{b}})Zhou, Jurafsky, and Hashimoto}]{zhou-etal-2023-navigating}
Kaitlyn Zhou, Dan Jurafsky, and Tatsunori Hashimoto. 2023{\natexlab{b}}.
\newblock Navigating the grey area: How expressions of uncertainty and overconfidence affect language models.
\newblock In \emph{Proceedings of the 2023 Conference on Empirical Methods in Natural Language Processing}, pages 5506--5524, Singapore. Association for Computational Linguistics.

\end{thebibliography}
\newpage
\appendix

\section{Analysis (cont.)}
\label{appendix:analysis_cont}

\begin{figure*}
    \centering
    \includegraphics[width=1\linewidth]{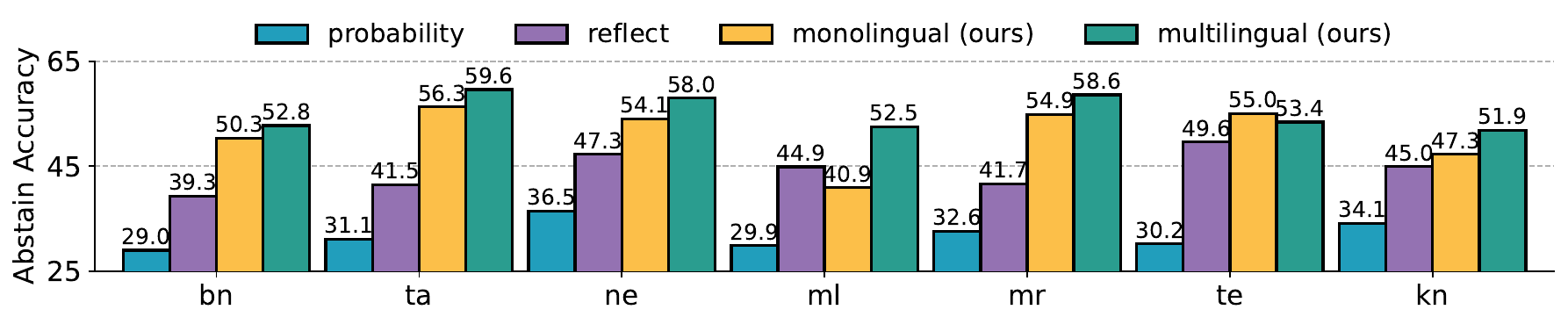}
    \caption{Abstain accuracy in the cross-lingual retrieval setting, where English Wikipedia is employed for retrieval to aid QA in low-resource languages. Multilingual feedback consistently produces more accurate abstain decisions in six of the seven low-resource languages.}
    \label{fig:retrieval}
\end{figure*}

\paragraph{\textsc{Multi-related} helps abstaining in cross-lingual retrieval.} When retrieval corpora are not readily available in low-resource languages, cross-lingual retrieval \citep{asai2021one, shen-etal-2022-recovering, huang2023improving, wieting-etal-2023-beyond, lin-etal-2023-maggretriever, thakur-etal-2024-leveraging} is often necessary for retrieval-augmented LLMs \citep{lewis2020retrieval, shi-etal-2024-replug, yasunaga2023retrieval, xu2024recomp}, where user queries are translated to high-resource languages and retrieval is performed with that language. We investigate whether our multilingual feedback approach works in this setting: we use English Wikipedia for retrieval \footnote{We retrieve Wikipedia with the WikiSearch API.} and prepend back-translated paragraphs before the query from the seven low-resource languages. We evaluate various abstain approaches with \textsc{ChatGPT} and present performance in Figure \ref{fig:retrieval}. Our proposed \emph{multilingual feedback} approach outperforms baselines for six of the seven low-resource languages, by 6.9\% on average. This indicates that our \emph{Multi-related} approach could also improve multilingual LLM reliability in retrieval-augmented settings.

\begin{figure}
    \centering
    \includegraphics[width=1\linewidth]{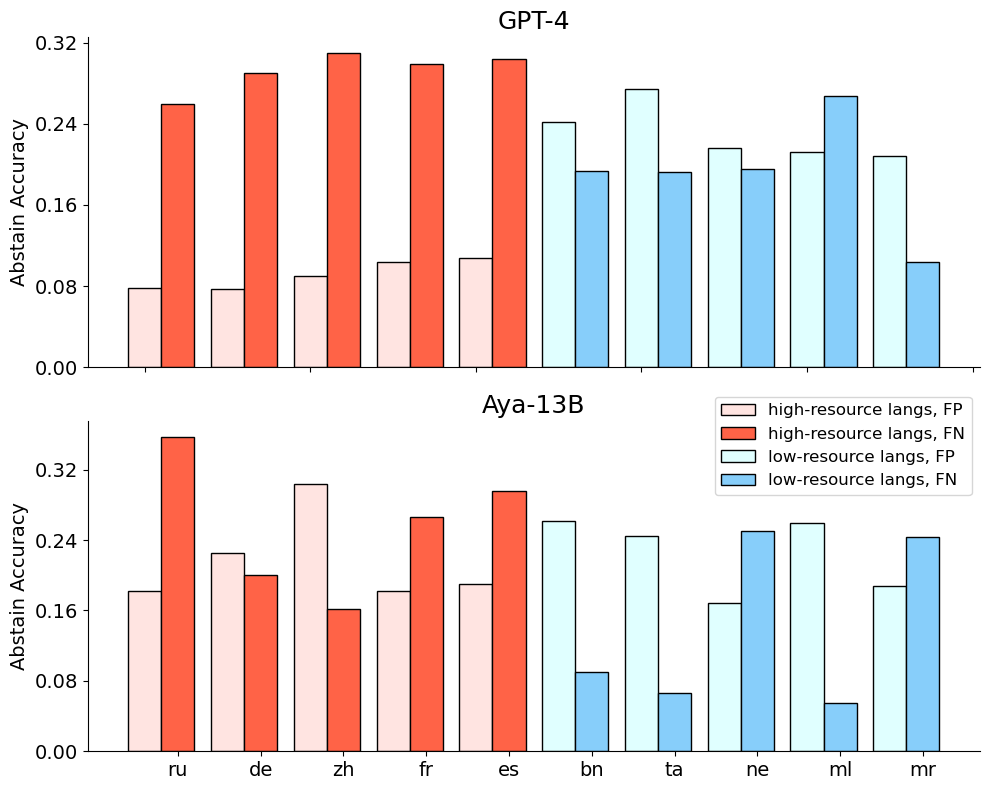}
    \caption{False positives and false negatives for \textsc{Multi-related} across low- and high-resource languages.}
    \label{fig:fpfn}
\end{figure}

\paragraph{FP and FN} False positives refer to cases where the LLM should be able to provide the correct answer but abstained, while false negatives are cases where the LLM did not abstain but generated an incorrect answer. We present the false positive and false negative rates of \textsc{Multi-related} in Figure \ref{fig:fpfn}: we find that on high-resource languages, LLMs tend to be more ``confident'' and the FN is usually higher; for low-resource languages, LLMs tend to be more ``conservative'' and the FP is usually higher. We argue that having a high FP for low-resource languages is desirable since LLM has diminishing factuality on the long tail of languages, thus LLMs should be more cautious and abstain more.

\begin{table}[t]
    \centering
    \resizebox{0.8\linewidth}{!}{
    \begin{tabular}{lccc}
         \toprule[1.5pt]
         \textbf{Method} & \textbf{Avg-H} & \textbf{Avg-M} & \textbf{Avg-L} \\ \midrule[0.75pt]
        \textsc{Probs} &0.8028 &\underline{0.7550} &0.3604 \\
        \textsc{Temp} &0.5988 &0.5583 &0.4331 \\
        \textsc{Ask Cali.} &0.4370 &0.4336 &\underline{0.6163} \\
        \textsc{Instruct} &\underline{0.8036} &0.7534 &0.3704 \\
        \textsc{Reflect} &0.5814 &0.5231 &0.4429 \\
        \textsc{MoreInfo} &0.7888 &0.7430 &0.3814 \\
        \textsc{BackTrans.} &0.6711 &0.6366 &0.4396 \\
        \textsc{SCthres.} &\bf 0.8136 &\bf 0.7688 &0.4634 \\
        \textsc{Conflict} &0.7072 &0.6926 &0.5741 \\ \midrule[0.75pt]
        \textsc{Mono-native} &0.6955 &0.6774 &0.4784 \\
        \textsc{Mono-English} &0.6048 &0.5891 &0.6009 \\
        \textsc{Multi-random} &0.7161 &0.6807 &0.4804 \\
        \textsc{Multi-related} &0.7906 &0.7445 &\bf 0.6386 \\ 
         \bottomrule[1.5pt]
    \end{tabular}
    }
    \caption{Performance averages for high, mid, and low-resource languages on Belebele (Avg-H, -M, and -L).}
    \label{tab:belebele}
\end{table}

\begin{figure}
    \centering
    \includegraphics[width=1\linewidth]{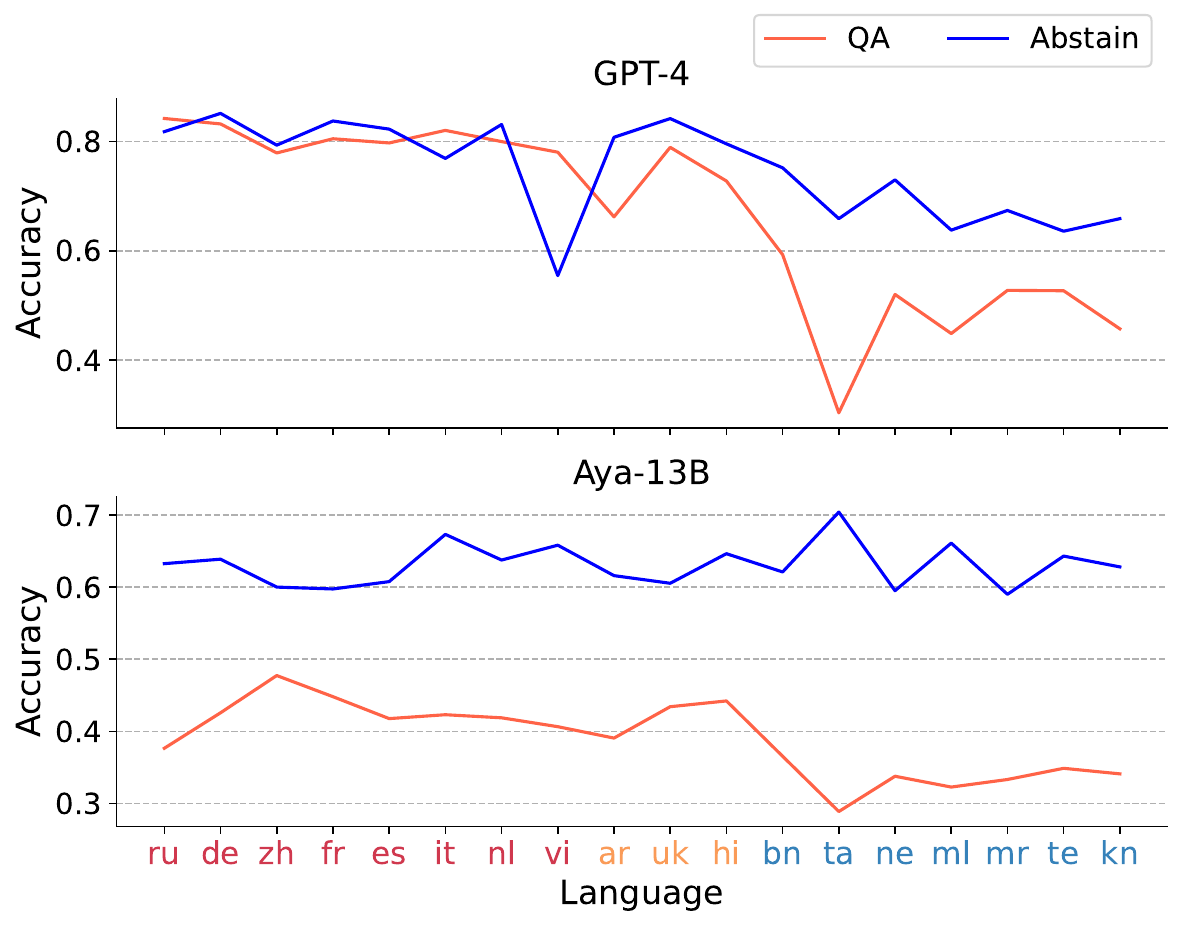}
    \caption{QA accuracy and abstain accuracy on the M-MMLU dataset with two LLMs. There is no lockstep synchronization between the two performance metrics, indicating that abstention is an independent research question. High-, mid-, and low-resource languages are labeled in red, orange, and blue colors, respectively.}
    \label{fig:correlation}
\end{figure}

\paragraph{Correlation between QA Performance and Abstain Performance} We present the question answering accuracy as well as the abstain accuracy across various languages in Figure \ref{fig:correlation}. We find that there is no lock-step synchronization between the two metrics, indicating that abstaining is an independent problem to question answering that needs further studies.

\paragraph{Another Dataset: Belebele} Belebele \citep{bandarkar2023belebele} is a multilingual reading comprehensive benchmark featuring parallel questions across 122 languages and variants. We evaluate baselines and our feedback-based approaches on Belebele and present the results in Table \ref{tab:belebele}. \textsc{Multi-related} achieves the best performance on low-resource language, while falling behind the strongest baselines in this reading comprehension setting. This motivates using different methodologies for abstention in different language contexts.

\paragraph{Working Examples} We conduct qualitative analysis to validate the generated feedback and abstain decisions. We specifically present several working examples in Tables \ref{tab:working_example_one}, \ref{tab:working_example_two}, and \ref{tab:working_example_three}.

\paragraph{Standard Deviation} Since \textsc{Multi-related} samples feedback from multiple languages, this sampling introduces randomness in the feedback content and potentially different abstain decisions. We re-run \textsc{Multi-related} three times with temperature $\tau = 0.7$, and we find that the standard deviation across runs is 0.0227, 0.0198, and 0.0086 for high, mid, and low-resource languages, indicating that the abstain performance is largely stable.

\paragraph{AbstainECE} Aside from a binary decision of abstaining or answering, the probabilities of the abstain decision token (True/False) could be employed as an indicator for probabilistic abstention. We present the AbstainECE metric \citep{feng2024don} in Table \ref{tab:abstainece}, which demonstrates that \textsc{Mono-English} and \textsc{Multi-related} are stronger while the latter is best for low-resource languages. We envision improving LLM calibration with multilingual contexts could also help.

\begin{table}[t]
    \centering
    \resizebox{0.8\linewidth}{!}{
    \begin{tabular}{lccc}
         \toprule[1.5pt]
         \textbf{Method} & \textbf{Avg-H} & \textbf{Avg-M} & \textbf{Avg-L} \\ \midrule[0.75pt]
        \textsc{Mono-native} &0.4594 &0.4630 &0.4276 \\
        \textsc{Mono-English} &\bf 0.4410 &\bf 0.4314 &\underline{0.4114} \\
        \textsc{Multi-random} &0.4713 &0.4829 &0.4475 \\
        \textsc{Multi-related} &\underline{0.4426} &\underline{0.4476} &\bf 0.3990 \\
         \bottomrule[1.5pt]
    \end{tabular}
    }
    \caption{AbstainECE averages for high, mid, and low-resource languages on Belebele (Avg-H, -M, and -L), the lower the better.}
    \label{tab:abstainece}
\end{table}

\begin{table}[t]
    \centering
    \resizebox{1\linewidth}{!}{
    \begin{tabular}{lccccc}\toprule[1.5pt]
    &1 &2 &3 &4 &5 \\\midrule[0.75pt]
    Avg-H &0.5768 &0.5878 &0.5397 &0.5856 &0.5703 \\
    Avg-M &0.5484 &0.5528 &0.5435 &0.5556 &0.5501 \\
    Avg-L &0.4688 &0.4959 &0.5825 &0.5003 &0.5004 \\
    \bottomrule[1.5pt]
    \end{tabular}
    }
    \caption{Abstain accuracy with one to five feedback(s) with \textsc{Multi-random}, \textsc{Aya-13B}, and M-MMLU.}
    \label{tab:feedback_number}
\end{table}

\begin{table}[!htp]\centering
\scriptsize
\resizebox{1\linewidth}{!}{
\begin{tabular}{lcccc}\toprule
&high &mid &low &equity ($\downarrow$) \\\midrule
default &\bf 0.6411 &\bf 0.5861 &\bf 0.4432 &0.0943 \\
reverse &0.6285 &0.5297 &0.4184 &0.1294 \\
self-included &0.6055 &0.5638 &0.4429 &\bf 0.0743 \\
lang var. &0.5790 &0.4488 &0.3726 &0.0964 \\
\bottomrule
\end{tabular}
}
\caption{Ablation settings of \textsc{Multi-related}, with ChatGPT and M-MMLU.}
\label{tab:ablation_settings}
\end{table}

\begin{table*}[!htp]\centering
\scriptsize
\resizebox{0.8\linewidth}{!}{
\begin{tabular}{lcccccc}\toprule[1.5pt]
&\multicolumn{3}{c}{Reliable Accuracy} &\multicolumn{3}{c}{Effective Reliability} \\\cmidrule{2-7}
&Avg-H &Avg-M &Avg-L &Avg-H &Avg-M &Avg-L \\\midrule[0.75pt]
\textsc{Probs} &0.4772 &0.4800 &0.3868 &-0.0269 &-0.0232 &-0.1220 \\
\textsc{Temp} &0.4635 &0.4651 &0.3694 &-0.0472 &-0.0442 &-0.1705 \\
\textsc{Ask Cal.} &0.5297 &0.5207 &0.4012 &0.0209 &0.0143 &-0.1069 \\
\textsc{Instruct} &0.4255 &0.4256 &0.3477 &-0.1412 &-0.1404 &-0.2538 \\
\textsc{Reflect} &0.4069 &0.4019 &0.3364 &-0.1843 &-0.1944 &-0.3260 \\
\textsc{BackTrans.} &0.4277 &0.4198 &0.3517 &-0.1292 &-0.1445 &-0.2648 \\
\textsc{SCThres.} &0.5389 &\textbf{0.5254} &0.3878 &0.0260 &\textbf{0.0172} &-0.0547 \\
\textsc{Conflict} &0.4585 &0.4559 &0.3636 &-0.0316 &-0.0338 &-0.1004 \\ \midrule[0.75pt]
\textsc{Mono-native} &0.4333 &0.4437 &0.3973 &-0.0705 &-0.0604 &-0.0821 \\
\textsc{Mono-English} &0.4796 &0.4594 &0.3884 &-0.0242 &-0.0406 &-0.0695 \\
\textsc{Multi-random} &0.4565 &0.4376 &0.3640 &-0.0344 &-0.0416 &-0.0656 \\
\textsc{Multi-related} &\textbf{0.5402} &0.4973 &\textbf{0.4474} &\textbf{0.0279} &-0.0077 &\textbf{-0.0289} \\
\bottomrule[1.5pt]
\end{tabular}
}
\caption{Other AbstainQA metrics with \textsc{Aya-13B} and MMLU.}
\label{tab:other_metrics}
\end{table*}

\paragraph{Number of feedback} We employ from one to five pieces of feedback in \textsc{Multi-random} and report performance in Table \ref{tab:feedback_number}. We find that performance saturates with three pieces of feedback, while it also fluctuates across language resourceness levels.

\paragraph{Other Metrics} In addition to abstain accuracy, we additionally report two more AbstainQA metrics, reliable accuracy \citep{feng2024don} and effective reliability \citep{whitehead2022reliable, si2023getting} in Table \ref{tab:other_metrics}. \textsc{Multi-related} achieves the best performance in four of the six settings.

\paragraph{Ablation Settings} We present several ablation setting of \textsc{Multi-related}: 1) \emph{reverse}, where the most distant languages are selected for feedback generation; 2) \emph{self-included}, where the language of the question is also employed to generate feedback; 3) \emph{lang var.}, where the feedback content stays the same but translated to different related languages. Table \ref{tab:ablation_settings} demonstrates that the default setting often works best, while including the original language for feedback generation could be beneficial for certain cases.

\paragraph{Another Interpretation of Abstain Overlap} For Figure \ref{fig:overlap}, another way is to compare the proportion of consensus, where LLMs abstain for zero or all three of languages. In this definition, the same conclusion still holds: the first control group has 23.1\% vs. 20.9\%, while the second control group has 32.1\% vs. 16.2\%.

\paragraph{Randomness in Sampling Feedback} We randomly sample feedback sets with temperature of 1 and repeat for 3 runs. If the LLM abstains/answers in all 3 runs, then it is deemed consistent; 1:2 and 2:1 scenarios are then deemed as inconsistent. We present results in Table \ref{tab:randomness}, showing that learning to abstain from multilingual feedback is largely consistent.

%Please add the following packages if necessary:
%\usepackage{booktabs, multirow} % for borders and merged ranges
%\usepackage{soul}% for underlines
%\usepackage{xcolor,colortbl} % for cell colors
%\usepackage{changepage,threeparttable} % for wide tables
%If the table is too wide, replace \begin{table}[!htp]...\end{table} with
%\begin{adjustwidth}{-2.5 cm}{-2.5 cm}\centering\begin{threeparttable}[!htb]...\end{threeparttable}\end{adjustwidth}
\begin{table}[!htp]\centering
\scriptsize
\resizebox{1\linewidth}{!}{
\begin{tabular}{lcccccc}\toprule[1.5pt]
\textbf{} &\textbf{bn} &\textbf{ml} &\textbf{mr} &\textbf{ne} &\textbf{ta} &\textbf{te} \\\midrule
MMLU, consistent &103 &101 &99 &117 &114 &95 \\
MMLU, inconsistent &42 &26 &35 &31 &21 &34 \\
consistent rate \% &71.03 &79.53 &73.88 &79.05 &84.44 &73.64 \\
Hellaswag, consistent &115 &136 &115 &131 &122 &122 \\
Hellaswag, inconsistent &39 &14 &39 &24 &18 &25 \\
consistent rate \% &74.68 &90.67 &74.68 &84.52 &87.14 &82.99 \\
\bottomrule[1.5pt]
\end{tabular}
}
\caption{Consistency when repeating the feedback sampling for three times.}
\label{tab:randomness}
\end{table}

\section{Experiment Details}
\label{appendix:experiment_details}

\paragraph{Dataset Details}
We employ M-MMLU, M-Hellaswag \citep{lai2023okapi}, and Belebele \citep{bandarkar2023belebele} as evaluations of multilingual AbstainQA. Specifically, we adopt 26 languages beyond English: 8 high-resource (Russian, ru; German, de; Chinese, zh; French, fr; Spanish, es; Italian, it; Dutch, nl; Vietnamese, vi), 11 mid-resource languages (Indonesian, id; Arabic, ar; Hungarian, hu; Romanian, ro; Danish, da; Slovak, sk; Ukrainian, uk; Catalan, ca; Serbian, sr; Croatian, hr; Hindi, hi), and 7 low-resource languages (Bengali, bn; Tamil, ta; Nepali, ne; Malayalam, ml; Marathi, Mr; Telugu, te; Kannada, kn). We follow the definition of language resourceness based on pretraining data frequency in \citet{lai2023okapi}. We randomly sample 200 questions for validation and 800 questions for test from each language, with minor variation across languages based on data availability.

\paragraph{Model Details} We employ the ``CohereForAI/aya-101'' model checkpoint on Huggingface for \textsc{Aya-13B}, and the Azure OpenAI API checkpoint of ``gpt4'' for \textsc{GPT-4}, and the ``GPT-3.5-TURBO-INSTRUCT'' model checkpoint on OpenAI API for \textsc{ChatGPT}.

\paragraph{Baseline Details} We refer readers to \citet{feng2024don} for a complete description of baselines. For the additional \textsc{BackTranslation} baseline, we translate the question to English and make an abstain decision in English, then use that abstain decision for other languages.

\paragraph{GPT-4 Evaluation Details}
For quality evaluation, we employ \emph{``Question: <question> Proposed Answer: <answer> Feedback 1: <feedback> Feedback 2: <feedback> Which feedback is more relevant to the question?''} and \emph{``Question: <question> Proposed Answer: <answer> Feedback 1: <feedback> Feedback 2: <feedback> Which feedback is more informative?''}. For role evaluation, we employ \emph{``Question: <question> Proposed Answer: <answer> Feedback 1: <feedback> Feedback 2: <feedback> Feedback 3: <feedback >What is the relationship among the three feedbacks? A. similar B. complementary C. conflicting D. unrelated Relationship:''}.

\paragraph{Implementation Details} We present the related languages employed for feedback generation in the Language Relatedness study (\Sref{sec:analysis}) in Tables \ref{tab:related_language_settings1}, \ref{tab:related_language_settings2}, and \ref{tab:related_language_settings3}.

\begin{table*}[ht]
\begin{tabularx}{\textwidth}{m{15.6cm}}
\toprule[1.5pt]
default:\{
            "en": ["German", "Dutch", "French"],
            "ru": ["Ukrainian", "Romanian", "Catalan"],
            "de": ["Dutch", "English", "French"],
            "zh": ["Arabic", "Slovak", "Danish"],
            "fr": ["Catalan", "German", "Spanish"],
            "es": ["Catalan", "Romanian", "French"],
            "it": ["Catalan", "Romanian", "Ukrainian"],
            "nl": ["German", "Italian", "Ukrainian"],
            "vi": ["Indonesian", "English", "Bengali"],
            "id": ["Vietnamese", "Catalan", "Russian"],
            "ar": ["Chinese", "Slovak", "Danish"],
            "hu": ["Romanian", "German", "French"],
            "ro": ["Catalan", "Italian", "Spanish"],
            "da": ["Slovak", "Dutch", "Ukrainian"],
            "sk": ["Chinese", "Arabic", "Danish"],
            "uk": ["Russian", "Italian", "Croatian"],
            "ca": ["Romanian", "Spanish", "Italian"],
            "sr": ["Slovak", "Danish", "Croatian"],
            "hr": ["Ukrainian", "Italian", "Dutch"],
            "hi": ["Bengali", "Talugu", "Marathi"],
            "bn": ["Hindi", "Telugu", "Nepali"],
            "ta": ["Malayalam", "Marathi", "Kannada"],
            "ne": ["Kanaada", "Telugu", "Hindi"],
            "ml": ["Tamil", "Marathi", "Kannada"],
            "mr": ["Tamil", "Malayalam", "Hindi"],
            "te": ["Kannada", "Tamil", "Nepali"],
            "kn": ["Telugu", "Malaayalam", "Tamil"]
        \} \\ \midrule[0.75pt]
syntactic:{
        "en": ["Spanish", "German", "French"],
        "ru": ["Ukrainian", "German", "Spanish"],
        "de": ["Dutch", "English", "Russian"],
        "zh": ["Arabic", "Slovak", "Hungarian"],
        "fr": ["Spanish", "English", "German"],
        "es": ["English", "French", "Russian"],
        "it": ["Catalan", "Romanian", "Dutch"],
        "nl": ["German", "Italian", "Danish"],
        "vi": ["Indonesian", "English", "French"],
        "id": ["Vietnamese", "English", "Italian"],
        "ar": ["Chinese", "Slovak", "Hungarian"],
        "hu": ["Russian", "Italian", "Romanian"],
        "ro": ["Italian", "Ukrainian", "Spanish"],
        "da": ["Dutch", "German", "French"],
        "sk": ["Chinese", "Arabic", "Hungarian"],
        "uk": ["Russian", "Italian", "Romanian"],
        "ca": ["Italian", "Dutch", "Romanian"],
        "sr": ["Catalan", "Ukrainian", "German"],
        "hr": ["Serbian", "Vietnamese", "Danish"],
        "hi": ["Kannada", "Russian", "Ukrainian"],
        "bn": ["Marathi", "Hindi", "Tamil"],
        "ta": ["Telugu", "Kannada", "Marathi"],
        "ne": ["Kannada", "Telugu", "Hindi"],
        "ml": ["Telugu", "Kannada", "Tamil"],
        "mr": ["Tamil", "Bengali", "Telugu"],
        "te": ["Tamil", "Nepali", "Kannada"],
        "kn": ["Tamil", "Nepali", "Hindi"],
        } \\ \midrule[0.75pt]
featural:{
        "en": ["German", "Russian", "French"],
        "ru": ["Romanian", "Ukrainian", "English"],
        "de": ["English", "French", "Dutch"],
        "zh": ["Arabic", "Slovak", "English"],
        "fr": ["German", "English", "Russian"],
        "es": ["English", "Russian", "French"],
        "it": ["Dutch", "Romanian", "Ukrainian"],
        "nl": ["German", "Italian", "English"],
        "vi": ["Indonesian", "English", "French"],
        "id": ["Vietnamese", "Catalan", "English"],
        "ar": ["Chinese", "Slovak", "English"],
        "hu": ["Rominian", "English", "Russian"],
        "ro": ["Russian", "Italian", "Hungarian"],
        "da": ["Serbian", "English", "Russian"],
        "sk": ["Chinese", "Arabic", "English"],
        "uk": ["Russian", "Italian", "Romanian"],
        "ca": ["Italian", "Dutch", "Romanian"],
        "sr": ["Danish", "Russian", "Spanish"],
        "hr": ["Catalan", "English", "Russian"],
        "hi": ["Bengali", "Nepali", "Telugu"],
        "bn": ["Hindi", "Nepali", "Telugu"],
        "ta": ["Malayalam", "Marathi", "Telugu"],
        "ne": ["Hindi", "Bengali", "Marathi"],
        "ml": ["Tamil", "Marathi", "Kannada"],
        "mr": ["Tamil", "Nepali", "Malayalam"],
        "te": ["Hindi", "Bengali", "Tamil"],
        "kn": ["Hindi", "Tamil", "Nepali"],
        } \\ \midrule[0.75pt]
genetic:{
        "en": ["German", "Dutch", "Danish"],
        "ru": ["Ukrainian", "Slovak", "Serbian"],
        "de": ["Dutch", "English", "Danish"],
        "zh": ["English", "Russian", "German"],
        "fr": ["Spanish", "Catalan", "Italian"],
        "es": ["Catalan", "Romanian", "French"],
        "it": ["Romanian", "Catalan", "Spanish"],
        "nl": ["German", "English", "Danish"],
        "vi": ["English", "Russian", "German"],
        "id": ["English", "Russian", "German"],
        "ar": ["English", "Russian", "German"],
        "hu": ["English", "Russian", "German"],
        "ro": ["Spanish", "Italian", "Catalan"],
        "da": ["German", "English", "Dutch"],
        "sk": ["Russian", "Ukrainian", "Serbian"],
        "uk": ["Russian", "Slovak", "Serbian"],
        "ca": ["Spanish", "Romanian", "Italian"],
        "sr": ["Croatian", "Russian", "Ukrainian"],
        "hr": ["Serbian", "Russian", "Slovak"],
        "hi": ["Bengali", "Marathi", "German"],
        "bn": ["Hindi", "Marathi", "English"],
        "ta": ["Malayalam", "Kannada", "Telugu"],
        "ne": ["English", "Russian", "German"],
        "ml": ["Tamil", "Kannada", "Telugu"],
        "mr": ["Hindi", "Bengali", "Russian"],
        "te": ["Tamil", "Malayalam", "Kannada"],
        "kn": ["Malayalam", "Tamil", "Telugu"],
        } \\ 
        \bottomrule[1.5pt]
\end{tabularx}
\caption{Related languages across different method settings, part 1.}
\label{tab:related_language_settings1}
\end{table*}

\begin{table*}[ht]
\begin{tabularx}{\textwidth}{m{15.6cm}}
\toprule[1.5pt]
geographic:{
        "en": ["French", "Dutch", "Danish"],
        "ru": ["English", "German", "French"],
        "de": ["French", "Italian", "Dutch"],
        "zh": ["English", "Russian", "German"],
        "fr": ["English", "German", "Spanish"],
        "es": ["French", "Catalan", "English"],
        "it": ["German", "French", "Hungarian"],
        "nl": ["English", "German", "French"],
        "vi": ["Indonesian", "Bengali", "Nepali"],
        "id": ["Vietnamese", "Bengali", "Tamil"],
        "ar": ["English", "Russian", "German"],
        "hu": ["German", "Italian", "Romanian"],
        "ro": ["German", "Italian", "Hungarian"],
        "da": ["English", "German", "French"],
        "sk": ["German", "Italian", "Hungarian"],
        "uk": ["German", "Hungarian", "Romanian"],
        "ca": ["French", "Spanish", "Italian"],
        "sr": ["German", "Italian", "Hungarian"],
        "hr": ["German", "Italian", "Hungarian"],
        "hi": ["Nepali", "Marathi", "Telugu"],
        "bn": ["Nepali", "Vietnamese", "Hindi"],
        "ta": ["Malayalam", "Marathi", "Telugu"],
        "ne": ["Hindi", "Bengali", "Vietnamese"],
        "ml": ["Tamil", "Marathi", "Telugu"],
        "mr": ["Hindi", "Tamil", "Malayalam"],
        "te": ["Hindi", "Tamil", "Malayalam"],
        "kn": ["Tamil", "Malayalam", "Marathi"],
        } \\ \midrule[0.75pt]
inventory:{
        "en": ["German", "Marathi", "Telugu"],
        "ru": ["Ukrainian", "Croatian", "Romanian"],
        "de": ["Dutch", "French", "English"],
        "zh": ["Arabic", "Danish", "Slovak"],
        "fr": ["Hungarian", "Dutch", "German"],
        "es": ["Hungarian", "German", "Indonesian"],
        "it": ["Catalan", "Romanian", "Ukrainian"],
        "nl": ["German", "French", "Hungarian"],
        "vi": ["English", "Dutch", "German"],
        "id": ["Catalan", "Croatian", "Romanian"],
        "ar": ["Chinese", "Danish", "Slovak"],
        "hu": ["French", "Romanian", "Italian"],
        "ro": ["Ukranian", "Catalan", "Italian"],
        "da": ["Chinese", "Arabic", "Slovak"],
        "sk": ["Chinese", "Arabic", "Danish"],
        "uk": ["Romanian", "Russian", "Italian"],
        "ca": ["Indonesian", "Italian", "Romanian"],
        "sr": ["Chinese", "Arabic", "Danish"],
        "hr": ["Catalan", "Indonesian", "Hungarian"],
        "hi": ["Telugu", "Bengali", "Nepali"],
        "bn": ["Telugu", "Nepali", "Hindi"],
        "ta": ["Kannada", "Malayalam", "Marathi"],
        "ne": ["Marathi", "Bengali", "Kannada"],
        "ml": ["Kannada", "Marathi", "Tamil"],
        "mr": ["Kannada", "Malayalam", "Nepali"],
        "te": ["Hindi", "Bengali", "Nepali"],
        "kn": ["Malayalam", "Marathi", "Tamil"],
        } \\ \midrule[0.75pt]
phonological:{
        "en": ["Indonesian", "Russian", "Catalan"],
        "ru": ["Catalan", "Hungarian", "Hindi"],
        "de": ["French", "Hungarian", "English"],
        "zh": ["Italian", "Dutch", "Arabic"],
        "fr": ["German", "Hungarian", "Hindi"],
        "es": ["English", "Russian", "Catalan"],
        "it": ["Chinese", "Dutch", "Arabic"],
        "nl": ["Chinese", "Italian", "Arabic"],
        "vi": ["Indonesian", "English", "Russian"],
        "id": ["English", "Russian", "Catalan"],
        "ar": ["Chinese", "Italian", "Dutch"],
        "hu": ["Russian", "Catalan", "German"],
        "ro": ["Russian", "Catalan", "German"],
        "da": ["Chinese", "Italian", "Dutch"],
        "sk": ["Chinese", "Italian", "Dutch"],
        "uk": ["Chinese", "Italian", "Dutch"],
        "ca": ["Russian", "Hungarian", "Hindi"],
        "sr": ["Spanish", "Chinese", "Italian"],
        "hr": ["Chinese", "Italian", "Dutch"],
        "hi": ["Russian", "Catalan", "French"],
        "bn": ["Telugu", "Kannada", "Russian"],
        "ta": ["Chinese", "Italian", "Dutch"],
        "ne": ["Romanian", "Telugu", "Kannada"],
        "ml": ["Chinese", "Italian", "Dutch"],
        "mr": ["Chinese", "Italian", "Dutch"],
        "te": ["Kannada", "Russian", "Catalan"],
        "kn": ["Kannada", "Russian", "Catalan"],
        } \\ 
        \bottomrule[1.5pt]
\end{tabularx}
\caption{Related languages across different method settings, part 2.}
\label{tab:related_language_settings2}
\end{table*}

\begin{table*}[ht]
\begin{tabularx}{\textwidth}{m{15.6cm}}
\toprule[1.5pt]
WVS:{
        "en": ["English", "English", "English"],
        "ru": ["Ukrainian", "Romanian", "Russian"],
        "de": ["German", "Dutch", "Danish"],
        "zh": ["Chinese", "Chinese", "Chinese"],
        "fr": ["French", "Slovak", "Hungarian"],
        "es": ["French", "Slovak", "Hungarian"],
        "it": ["French", "Slovak", "Hungarian"],
        "nl": ["German", "Dutch", "Danish"],
        "vi": ["Vietnamese", "Vietnamese", "Vietnamese"],
        "id": ["Indonesian", "Indonesian", "Indonesian"],
        "ar": ["Arabic", "Hindi", "Bengali"],
        "hu": ["French", "Slovak", "Hungarian"],
        "ro": ["Ukrainian", "Romanian", "Russian"],
        "da": ["German", "Dutch", "Danish"],
        "sk": ["French", "Slovak", "Hungarian"],
        "uk": ["Ukrainian", "Romanian", "Russian"],
        "ca": ["Catalan", "Catalan", "Catalan"],
        "sr": ["Serbian", "Serbian", "Serbian"],
        "hr": ["French", "Slovak", "Hungarian"],
        "hi": ["Arabic", "Hindi", "Bengali"],
        "bn": ["Arabic", "Hindi", "Bengali"],
        "ta": ["Arabic", "Hindi", "Bengali"],
        "ne": ["Arabic", "Hindi", "Bengali"],
        "ml": ["Arabic", "Hindi", "Bengali"],
        "mr": ["Arabic", "Hindi", "Bengali"],
        "te": ["Arabic", "Hindi", "Bengali"],
        "kn": ["Arabic", "Hindi", "Bengali"],
        } \\ \midrule[0.75pt]
LLM-generated:{
        "en": ["Frisian", "Dutch", "German"],
        "ru": ["Belarusian", "Ukrainian", "Rusyn"],
        "de": ["Dutch", "Luxembourgish", "Yiddish"],
        "zh": ["Cantonese", "Shanghainese", "Hokkien"],
        "fr": ["Italian", "Spanish", "Catalan"],
        "es": ["Portuguese", "Catalan", "Italian"],
        "it": ["Sicilian", "Neapolitan", "Tuscan"],
        "nl": ["Afrikaans", "Frisian", "German"],
        "vi": ["Muong", "Khmer", "Mon"],
        "id": ["Malay", "Minangkabau", "Javanese"],
        "ar": ["Hebrew", "Aramaic", "Amharic"],
        "hu": ["Finnish", "Estonian", "Udmurt"],
        "ro": ["Italian", "Spanish", "French"],
        "da": ["Swedish", "Norwegian", "Icelandic"],
        "sk": ["Czech", "Polish", "Slovene"],
        "uk": ["Russian", "Belarusian", "Rusyn"],
        "ca": ["Occitan", "Spanish", "French"],
        "sr": ["Croatian", "Bosnian", "Montenegrin"],
        "hr": ["Serbian", "Bosnian", "Montenegrin"],
        "hi": ["Urdu", "Punjabi", "Bengali"],
        "bn": ["Assamese", "Odia", "Maithili"],
        "ta": ["Kannada", "Telugu", "Malayalam"],
        "ne": ["Maithili", "Bhojpuri", "Awadhi"],
        "ml": ["Tamil", "Tulu", "Kannada"],
        "mr": ["Konkani", "Sanskrit", "Gujarati"],
        "te": ["Kannada", "Tamil", "Malayalam"],
        "kn": ["Telugu", "Tamil", "Tulu"],
        } \\
        \bottomrule[1.5pt]
\end{tabularx}
\caption{Related languages across different method settings, part 3.}
\label{tab:related_language_settings3}
\end{table*}

\begin{table*}[ht]
\begin{tabularx}{\textwidth}{m{15.6cm}}
% {\textwidth}{l}
\toprule[1pt]
Question:
\begin{CJK*}{UTF8}{gbsn}
非营利组织活动报告表的主要目的是提供相关信息给其 \ A: 资源提供者 \ B: 管理者 \ C: 受益人 \ D: 州监管机构
\end{CJK*} \\
(translated: The main purpose of the non-profit organization activity report form is to provide relevant information to its A: resource providers. B: managers. C: beneficiaries. D: state regulatory agencies.) \\ \\
Proposed Answer: C \\ \\

Feedback 1: \\
\includegraphics[width=1\linewidth]{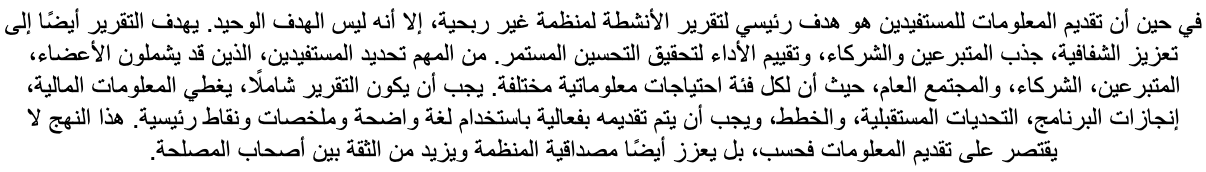}
\\
(translated: While providing information to beneficiaries is a key goal of a non-profit organization's activity report, \emph{it is not the sole purpose}. The report also aims to enhance transparency, attract donors and partners, and evaluate performance for continuous improvement. Specifying the beneficiaries, who may include members, donors, partners, and the general community, is essential as each group has different informational needs.) \\ \\

Feedback 2: \\
Okrem informovania prijímateľov správa často slúži aj na preukázanie transparentnosti organizácie voči donorom, regulačným orgánom, dobrovoľníkom a verejnosti. Pomáha tiež pri hodnotení efektívnosti a dosiahnutých výsledkov, čo môže podporiť ďalší rozvoj organizácie a jej projektov. Celkovo je teda cieľom správy nielen informovať, ale aj budovať dôveru a zodpovednosť voči všetkým zainteresovaným stranám. \\
(translated: In addition to informing beneficiaries, \emph{the report often serves to demonstrate the organization's transparency to donors}, regulatory bodies, volunteers, and the public. It also helps in evaluating the effectiveness and outcomes achieved, which can support the further development of the organization and its projects. Overall, the report's purpose is not only to inform but also to build trust and accountability towards all stakeholders involved.) \\ \\

Feedback 3: \\
\begin{CJK*}{UTF8}{gbsn}
非营利组织的主要目的之一是,向其受益人提供有关组织活动的反馈信息,主要依据组织的运作特征. 受益人包括项目组织的利益相关者,如基金管理人,或组织会员,如捐赠者。 非营利组织的负责人应该遵守对组织的监管规定。
\end{CJK*} \\
(translated: One of the main purposes of non-profit organizations is to provide feedback to their beneficiaries about the organization's activities, based primarily on the operational characteristics of the organization. Beneficiaries include stakeholders of project organizations, such as fund managers, or \emph{organizational members, such as donors}. The leaders of non-profit organizations should comply with regulatory requirements for the organization.) \\ \\
Abstain: Yes (correct answer is A) \\
\bottomrule[1pt]
\end{tabularx}
\caption{Working example one, where the three pieces of feedback unanimously point out the error in the answer.}
\label{tab:working_example_one}
\end{table*}

\begin{table*}[ht]
\begin{tabularx}{\textwidth}{m{15.6cm}}
% {\textwidth}{l}
\toprule[1pt]
Question:
\begin{CJK*}{UTF8}{gbsn}
以下哪个提供了有关地球核心、地幔和下地壳结构的大部分信息？ A: 测量地球磁场的强度和波动 B: 潜流岩的检测 C: 从深部钻孔中收集样本 D: 研究通过地球的地震波传播的速度和路径
\end{CJK*} \\
(translated: Which of the following provides most of the information about the structure of the Earth's core, mantle, and lower crust? A: Measuring the strength and fluctuations of the Earth's magnetic field B: Detecting plumes of molten rock C: Collecting samples from deep drilling D: Studying the speed and paths of seismic waves passing through the Earth) \\ \\
Proposed Answer: C \\ \\

Feedback 1: \\
\includegraphics[width=1\linewidth]{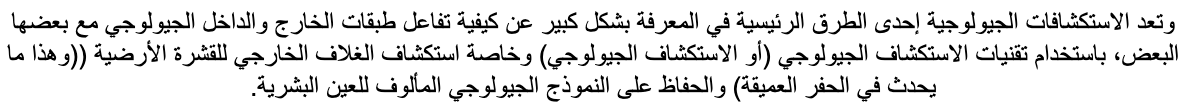}
\\
(translated: Geological exploration is considered one of the main ways to significantly increase our knowledge about how the layers of the Earth's exterior and interior interact with each other, using geological exploration techniques, especially exploring the outer shell of the Earth's crust (as in deep drilling) and maintaining the familiar geological model for the human eye.) \\ \\

Feedback 2: \\
Materiály z inžinierskej činnosti sa dokážu zaujímať o štruktúru zeme vďaka spôsobu, akým tvoria základne zemského plášťa, tiež o spôsobe používania jadrovej energie, uvoľňovania energie z dopadu gravitačných silov. Predpokladáme pôsobenie gravitácie na štruktúru centrálneho a periférneho plášťa, rovnako ako aj gravitačné účinky uvoľnenej energie medzi nimi. Na základe výsledkov dokážeme predpokladať, ako sa skladá zloženie povrchovej vrstvy, ako ovplyvňujú polárnu tepelnú uhľovodíkovú vrstvu, ako ovplyvňuje povrchovú vlhkosť, a ako ovplyvňuje teplotnú pozíciu rôznych vrstiev. \\
(translated: Materials from engineering activities can be interested in the structure of the earth due to the way they form the foundations of the earth's mantle, as well as the methods of using nuclear energy and releasing energy from the impact of gravitational forces. We assume the influence of gravity on the structure of the central and peripheral mantle, as well as the gravitational effects of the released energy between them. Based on the results, we can predict how the composition of the surface layer is formed, how it affects the polar thermal hydrocarbon layer, how it affects surface moisture, and how it affects the temperature position of different layers.) \\ \\

Feedback 3: \\
Undersøgelse af jordens magnetfelt og dens bevægelser, samt måling af jordoverfladens temperatur og fugtighed er de mest anvendte metoder. Undersøgelsen af jordens magnetfelt og dens bevægelser er ikke alene vigtigt for forståelsen af jordens magnetfelt, men også påvirker andre områder, såsom jordskælv og vulkaner, hvor man kan studere jordens magnetfelt, mens man i samme tid observere andre systemer, såsom atmosfære og havanlæg. \\
(translated: The investigation of the Earth's magnetic field and its movements, as well as the measurement of surface temperature and humidity, are the most commonly used methods. The study of the Earth's magnetic field and its movements is not only important for understanding the Earth's magnetic field, but also affects other areas, such as earthquakes and volcanoes, where one can study the Earth's magnetic field while observing other systems, such as the atmosphere and oceanic structures, at the same time.) \\ \\
Abstain: Yes (correct answer is D) \\
\bottomrule[1pt]
\end{tabularx}
\caption{Working example two, where there is a conflict among the three feedback.}
\label{tab:working_example_two}
\end{table*}

\begin{table*}[ht]
\begin{tabularx}{\textwidth}{m{15.6cm}}
% {\textwidth}{l}
\toprule[1pt]
Question:
\begin{CJK*}{UTF8}{gbsn}
关于总统提名大会代表，以下哪个陈述是正确的？ A: 代表通常更有可能在某个时候注册为第三方投票者。 B: 大多数普通人比代表更有教育背景。 C: 代表通常对政治没有太大兴趣。 D: 通常，普通人的意识形态比代表更少。
\end{CJK*} \\
(translated: Which statement is correct about delegates to the presidential nominating convention? A: Delegates are more likely to register as third-party voters at some point. B: Most ordinary people have a higher level of education than delegates. C: Delegates generally have less interest in politics. D: Typically, ordinary people have less ideology than delegates.) \\ \\
Proposed Answer: C \\ \\

Feedback 1: \\
\includegraphics[width=1\linewidth]{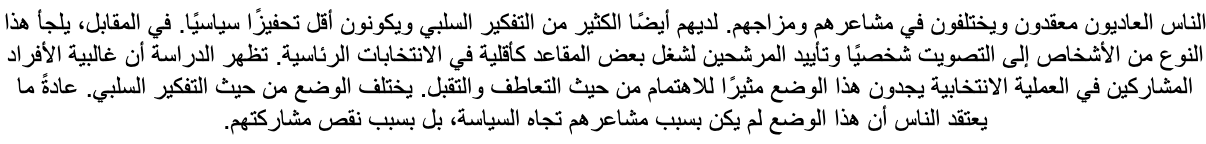}
\\
(translated: \emph{Ordinary people are complex and differ in their feelings and moods.} They also have a lot of negative thinking and are \emph{less politically motivated}. In contrast, this type of people resort to voting personally and endorse candidates to hold some seats as a minority in the presidential candidate. The study shows that the majority of individuals participating in the electoral process find this situation interesting in terms of empathy and embrace. It differs in terms of negative thinking. People usually believe that this situation was not due to their feelings towards politics, but due to their lack of participation.) \\ \\

Feedback 2: \\
Zvyčajne je obyčajný človek, ktorý podporuje kandidátov z celého štátu, zvyčajne ľuďom plne neznámym. Umožňujú voličom hlasovať za svojho kandidáta a predstavujú ho podľa svojho názoru. Keďže na zvolenie kandidáta je potrebný nejaký druh spoločného zvolenia, získali ďalší hlas členovia zvoliteľskej delegácie a títo sú usilujúci o účasť na zvolovaní. O takéto pôsobenie sa postará delegát zvoliteľskej delegácie, ktorý je ľudom plne neznámy. \\
(translated: Usually, an ordinary person who supports candidates from all over the state, usually to people completely unknown. They allow voters to vote for their candidate and represent him according to their opinion. Since some kind of common election is needed to elect a candidate, additional votes were obtained by members of the elector's delegation, who are seeking to participate in the election. A delegate of the elector's delegation, who is completely unknown to people, will take care of this action.) \\ \\

Feedback 3: \\
Delegaterne fra staten har ofte mere viden om politik end de fleste almindelige mennesker. \\
(translated: The state's delegates often have \emph{more knowledge about politics than most ordinary people}.) \\ \\
Abstain: Yes (correct answer is D) \\
\bottomrule[1pt]
\end{tabularx}
\caption{Working example three, where there is a conflict among the three feedback.}
\label{tab:working_example_three}
\end{table*}

\end{document}